\pdfoutput=1
\documentclass[11pt]{article}
\usepackage{graphicx}
\usepackage{multirow}
\usepackage{amsmath}
\usepackage{amsfonts}
\PassOptionsToPackage{hyphens}{url}\usepackage{hyperref}
\usepackage{ulem}
\usepackage{booktabs}
\usepackage{enumitem}
\usepackage[]{EMNLP2023}

\usepackage{times}
\usepackage{latexsym}

\usepackage[T1]{fontenc}
\usepackage[utf8]{inputenc}

\usepackage{microtype}

\usepackage{inconsolata}

\title{WSDMS: Debunk Fake News via \underline{W}eakly \underline{S}upervised \underline{D}etection of \underline{M}isinforming \underline{S}entences with Contextualized Social Wisdom}

\author{Ruichao Yang$^{1}$, Wei Gao$^{2}$, Jing Ma$^{1}$\thanks{\; Jing Ma is the corresponding author.}, Hongzhan Lin$^{1}$, Zhiwei Yang$^{3}$\\
$^{1}$Hong Kong Baptist University, Hong Kong SAR, China\\
$^{2}$Singapore Management University, Singapore\\
$^{3}$Jinan University, Guangzhou, Guangdong, China\\
\texttt{\{csrcyang,majing,cshzlin\}@comp.hkbu.edu.hk}, \\\texttt{weigao@smu.edu.sg}, \texttt{yangzw@jnu.edu.cn}
}

\begin{document}
\maketitle
\begin{abstract}
In recent years, we witness the explosion of false and unconfirmed information (i.e., rumors) that went viral on social media and shocked the public. Rumors can trigger versatile, mostly controversial stance expressions among social media users. Rumor verification and stance detection are different yet relevant tasks. Fake news debunking primarily focuses on determining the truthfulness of news articles, which oversimplifies the issue as fake news often combines elements of both truth and falsehood. Thus, it becomes crucial to identify specific instances of misinformation within the articles. In this research, we investigate a novel task in the field of fake news debunking, which involves detecting sentence-level misinformation. One of the major challenges in this task is the absence of a training dataset with sentence-level annotations regarding veracity. Inspired by the Multiple Instance Learning (MIL) approach, we propose a model called Weakly Supervised Detection of Misinforming Sentences (WSDMS). This model only requires bag-level labels for training but is capable of inferring both sentence-level misinformation and article-level veracity, aided by relevant social media conversations that are attentively contextualized with news sentences.
%Specifically, we implicitly correspond each sentence to conversation trees via cross-attention to facilitate spotting misinformed sentences and design a graph-based attention to bridge the sentence- and article-level labels. 
We evaluate WSDMS on three real-world benchmarks and demonstrate that it outperforms existing state-of-the-art baselines in debunking fake news at both the sentence and article levels.
\end{abstract}

\section{Introduction} \label{sec:intro}
%Social media facilitates the rapid propagation of fake news, but 
%have proven to be Petri dishes for misinformation as they transform the ways people retrieve information and facilitate rapid information sharing and retrieval~\cite{vosoughi2018spread, cheng2021causal}. Meanwhile, 
%people are often poor at efficiently identifying the misinformation which is well-edited and/or composed of factual contexts and falsehoods. %articles due to not all sentences in a news article are misinformation. 
%For example, there is a fake news article claiming that ``COVID-19 vaccines are used to inject with a microchip for monitoring people''\footnote{\url{https://www.bbc.com/news/52847648}}. Malicious actors deliberately tampered with active ingredients in vaccines and added misinformation, such as microchips, to create panic in the whole society. With the emerging volume of misinformation every day, there is a strong need to automatically spot finer-grained misinformation, which will facilitate timely and transparent fake news detection. %fact-checking systems.

%Misinformation is defined as messages where people's beliefs about factual matters are not supported by clear evidence and expert opinion~\cite{nyhan2010corrections}, e.g., myths, rumors, fake news, etc. Earlier methods for misinformation detection exploit the textual or visual content of the news articles with deep neural models~\cite{wang2020weak, qian2021hierarchical}.  

Misinformation, such as fake news, poses tremendous risks and threats to contemporary society. The detection of fake news entails various technical challenges~\cite{glockner2022missing}, and one of them is accurately identifying false elements within news articles. This challenge arises due to the blending of authentic and fabricated content by creators of fake news, thereby complicating the determination of overall veracity~\cite{solovev2022moral}. Such instances have been prevalent during the Covid-19 pandemic\footnote{\url{https://ahmedabadmirror.com/gujarat-plans-to-give-world-a-wonder-drug-to-battle-corona/76017951.html}. This article combines factual information about the historical use of cow urine in India's traditional medicine with false assertions that cow urine contains active ingredients capable of treating Covid-19 and has been used in hospitals in South Korea and China.}. %Furthermore, the advent of large language models like ChatGPT has significantly lowered the bar of manipulating content during text generation, further amplifying the challenges of fake news.

Fake news detection aims to determine the veracity of a given news article~\cite{shu2017fake}. Previous analysis has revealed that users often share comments and provide evidence about fake news on social media platforms~\cite{zubiaga2017towards}, which has led to a growing stream of research that leverages these social engagements, along with the content of news articles, to aid in fake news detection~\cite{pan2018content,shu2019defend,min2022divide}. This approach bears analogies to rumor detection, where the focus is on assessing as a specific statement rather than an entire news article~\cite{wu2015false,ma2018rumor,bian2020rumor,lin2021rumor,song2021adversary,park2021experimental,zheng2022mfan,xu2022evidence}. Many studies in this domain aims to train supervised classifiers using features extracted from the social context and the content of the claim or article.  % \textcolor{blue}{Fake news is defined as false or misleading information that presented as news, aiming at achieving some ulterior motives~\cite{shu2017fake}. Existing methods for fake news detection usually find evidences from related reports or articles~\cite{popat2018declare,ma2019sentence}. However, fake news is usually deliberately created by distorting partial contents, which makes it nontrivial to detect solely relied on the articles/reports content yet. So recently, additional knowledge and social media information has been used to assist fake news detection~\cite{pan2018content,shu2019defend}, this is because some opinions and evidences will be shared through social media among users. Furthermore, some methods incorporate structural features mined from social media data with kernel learning~\cite{wu2015false,rosenfeld2020kernel}, recursive neural networks~\cite{ma2018rumor} or Graph Neural Networks~\cite{zhou2019network, fu2021sdg} has been adopted to mitigate the disseminating of fake news.}
%Social interactions among users targeting fake news can generate a large amounts of chaotic, incomplete and noisy data~\cite{tang2015line} yet, 
However, the existing fake news detection models predominately focus on coarse-level classification of the entire article, which oversimplifies the problem. Misinformation can be strategically embedded within an article by manipulating portions of its content to enhance its credibility~\cite{feng2012syntactic,rogers2017artful,zhu2022generalizing}
%We hypothesize that when these sentences are spotted, the veracity of the coarse-level article can be quickly determined. %At the same time, sentence-level misinformation detection also provides fact checkers and readers with more intuitive and powerful insights to judge the veracity of articles, saving manpower and time. 
%address the sentence fine-grained level misinformation detection problem in a unified way, 
Therefore, we target a fine-grained task that aims to identify sentences containing misinformation within an article, which can be jointly learned with article-level fake news detection. %\textcolor{blue}{claim/misinformation?} detection problem.
%simultaneously spot sentence-level misinformation and coarse-level article verification. Furthermore, our research can also be well-generalized on the traditional misinformation detection problem, owing to the entire article may be judged misinformation on the condition that some parts of the article have been deliberately distorted.
%While existing works have shown the importance of deep learning methods and modeling the propagation structure for misinformation detection, it is difficult for them to provide a clear and accurate understanding about the key sentences' veracity (i.e., misinformation sentence in a fake news article). In addition to accuracy, sentence-level detection also provides the capability to enhance fine-grained explainability in misinformation detection process. 

%\textcolor{cyan}{Comment: These two paragraphs are unnecessarily long. The argument should start from fake news detection rather than such a broad context of rumor claim detection and fact checking. Fake news detection is well established. Readers won't be confused by this work if such broad context is not given. Instead, arguing from such a broad context will confuse readers, as sentence-level misinformation in an article is closely related to fake news detection more than rumor detection and fact checking. Speaking how rumor detection in social media is relevant in the related work would suffice. Actually the abstract has argued for the work nicely. The introduction can focus on the motivating example and explain why social media can help in this task, which is more critical}.

\begin{figure*}[htbp]
\setlength{\belowcaptionskip}{-0.6cm}
\centering
% \vspace{-0.3cm}
\includegraphics[width=6in]{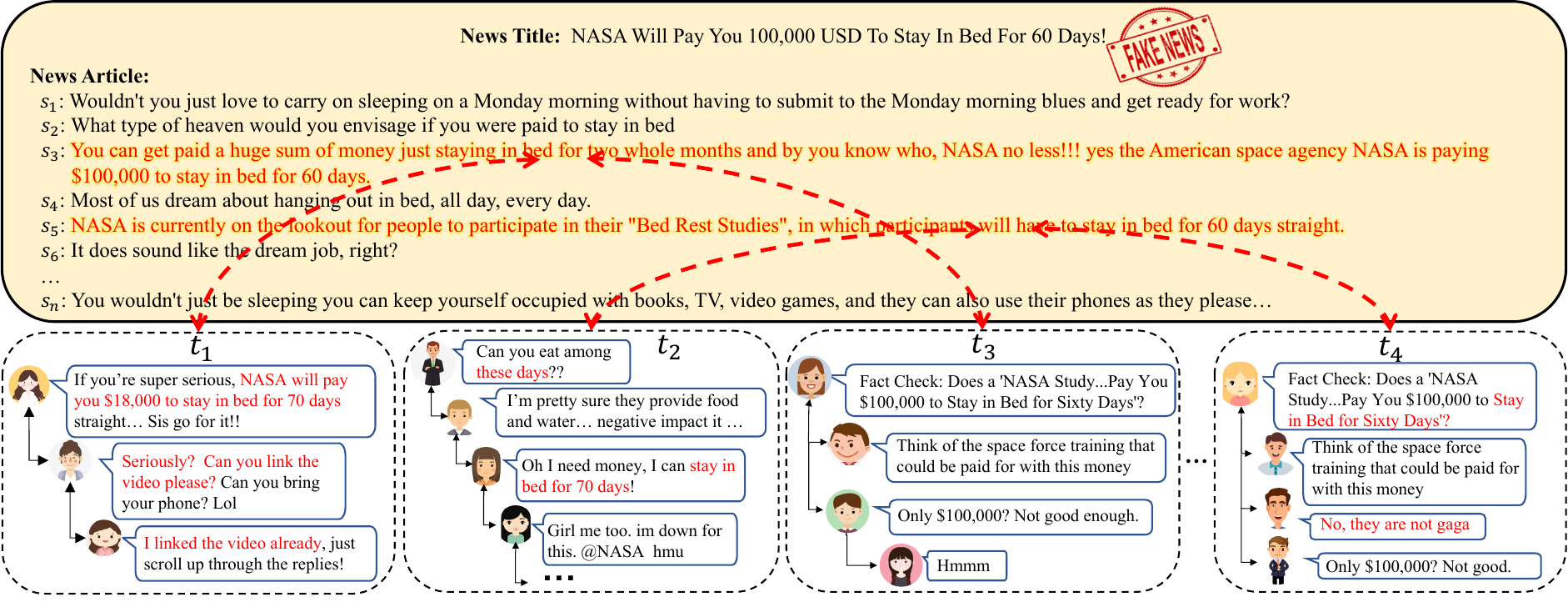}
\vspace{-0.1cm}
\caption{A fake news article together with its relevant social context information, where the sentences containing misinformation (i.e., $s_3$ and $s_5$) are in orange and the posts implying the misinforming sentences are in red.}%textcolor{red}{In the figure S should be s. Why there is '...' between S2 and S3? Pay attention to detail.}
\label{fig:intro}
\end{figure*}

Figure~\ref{fig:intro} shows an illustrative example of a fake news article titled ``NASA will pay 100,000 USD to participants staying in bed for 60 days!'', where the sentences in the article can be linked to a set of social conversations organized as propagation trees of posts. %thread of posts is represented as a propagation tree.
% Figure 1 exemplifies a fake news article about ``NASA will pay you to stay inn bed for 60 days!", where each sentence corresponds to a set of tweets and the tweets are constructed with conversation tree structure. %thread of posts is represented as a propagation tree. 
These sentences contain opinions and evidence that can aid in the veracity classification at the sentence and article levels, specifically in spotting misinformation sentences. For instance, sentence $s_3$ can be debunked by referring to trees $t_1$ and $t_3$, as they provide evidence that contradicts the incorrect reward amount and duration mentioned in the sentence. This information helps in determining that the article is fake. Conversely, if we already know that the article is fake, we can infer that there must be misinforming sentences present within it. 

%To spot misinformation, 
However, existing methods are not readily applicable for the identification of sentence-level misinformation due to two main reasons: 1) Obtaining veracity labels for sentences in an article is costly, as it requires annotators to exhaustively fact-check each sentence. 2) While rumor detection models can predict the label of a given claim, they often assume the availability of social conversations that correspond to the claim. However, it is difficult to establish a correspondence between social conversations and specific sentences within a news article. 
%Due to the distinct correlations between the sentence-level and coarse-level veracity, we assume that an article is predicted as fake news on the condition that at least one sentence contains misinformation. 
Inspired by multiple instance learning (MIL)~\cite{foulds2010review}, we attempt to develop an approach for debunking fake news via  \underline{w}eakly \underline{s}upervised \underline{d}etection of \underline{m}isinforming \underline{s}entences (i.e., instances), called WSDMS\footnote{\url{https://github.com/HKBUNLP/WSDMS-EMNLP2023}},
%\footnote{\url{https://anonymous.4open.science/r/WSDMS-55C5/}}
only using available article-level veracity annotations (i.e., bag-level labels) and a handful of social conversations related to the news.   

%We extend standard MIL and propose a novel \underline{h}ierarchical c\underline{o}arse-to-fine weakly supervised framework for spotting sen\underline{t}ence-level misinformation and determining the arti\underline{c}le-level f\underline{ake} news (\textbf{HotCake}\footnote{\url{https://anonymous.4open.science/r/HotCake-8F56/}}) simultaneously, which only requires coarse-level labels for model training. 
To gather the relevant social conversations associated with an article, we employ established methods used in fake news detection that rely on social news engagement data collection~\cite{shu2020fakenewsnet}, which provides the necessary conversation trees linked to the article in question. We devise a hierarchical embedding model to establish connections between each sentence in the article and its corresponding conversations, facilitating the identification of sentence-level misinformation. Standard MIL determines the bag-level label as positive if one or more instances within the bag are positive, and negative otherwise~\cite{dietterich1997solving}. To improve its tolerance on sentence-level prediction errors, we further develop a collective attention mechanism for a more accurate article veracity inference on top of the sentence-level predictions. The entire framework is trained end-to-end by optimizing a loss function that aims to alleviate prediction bias by considering both sentence- and article-level consistencies. Our approach ensures that the model captures the nuances of misinformation at both levels of granularity.
 %train our model effectively, which meanwhile alleviates the prediction bias at the sentence level. 
%We collect a new open-domain dataset from both news and social media outlets, and experimental results conducted on three real-world benchmarks demonstrate that our method surpasses state-of-the-art methods and gains performance on both sentence-level misinformation spotting and coarse-level article detection tasks at a large margin. %We will make all the source codes in our experiments publicly accessible.  
Our contributions are summarized as follows: %in three folds:
\begin{itemize}
\item 
Unlike existing fake news detection approaches, we introduce a new task that is focused on spotting misinforming sentences in news articles while simultaneously detecting article-level fake news.
\item 
We develop WSDMS, a MIL-based model, to contextualize news sentences aided by social conversations about the news and use only article veracity annotations to weakly supervise sentence representation and model training. 
\item 
Our method achieves superior performance over state-of-the-art baselines on sentence- and article-level misinformation detection.
\end{itemize}

%\textcolor{cyan}{Need to summarize the main contributions.}

\section{Related Work}

Early studies on fake news detection have attempted to exploit various approaches to extract features from news content and social context information, including linguistic features~\cite{potthast2018stylometric,azevedo2021lux}, visual clues~\cite{jin2016novel}, 
temporal traits~\cite{kwon2013prominent,ma2015detect}, 
user behaviors and profiles~\cite{castillo2011information,ruchansky2017csi,shu2019role}. Subsequent studies have employed neural networks to automatically learn deep feature representations from similar sources of data~\cite{ma2016detecting,popat2018declare,ma2019sentence,nguyen2020BERTweet,kaliyar2021fakebert, sheng2022zoom}. 
%that attend over evidential keywords~\cite{popat2018declare} and sentence-level evidence~\cite{ma2019sentence}.
%For example, \citet{popat2018declare} proposes a deep learning method to attend over evidential keywords, \citet{ma2019sentence} adopts hierarchical attention networks to attend on sentence-level evidence, 
%\citet{nguyen2020BERTweet} introduces a pre-trained Bert model with enormous tweet comments,
%\citet{kaliyar2021fakebert} develops a CNN-Bert model to improve the accuracy of fake news detection, 
Furthermore, researchers have incorporated  external knowledge sources~\cite{pan2018content,dun2021kan,hu2021compare} and combined multi-modal data~\cite{wang2018eann,wang2021multimodal,fung2021infosurgeon,wu2021multimodal,silva2021embracing,chen2022cross} to enhance learning and improve fake news detection performance.
%After that, external evidence~\cite{popat2018declare}, prior knowledge~\cite{pan2018content,dun2021kan,hu2021compare}, and multi-modal methods~\cite{wang2018eann,wang2021multimodal} have been used in subsequent studies, demonstarting the effectiveness of deep neural networks in misinformation detection tasks.
%\textcolor{blue}{external evidence}\citet{popat2018declare} proposed a neural network model to aggregate signals from external evidence articles and the source trustworthiness. 
%The \textcolor{blue}{multi-modal method} has also been used in subsequent studies~\cite{wang2018eann,wang2021multimodal}.
%\textcolor{red}{what is the entities? which one in the above utilize prior knowledge?} %But the entities in news content may be not comprehensive enough due to a lack of prior knowledge. To further enhance the understanding of the entities in the news content, 
%\citet{pan2018content,dun2021kan,hu2021compare} incorporate \textcolor{blue}{prior knowledge into fake news detection}. 
%However, these methods generally heavily depend on rich lexicons and knowledge bases, which require tedious and expensive manual effort. 
Notably, social context information has played a crucial role in debunking fake news and rumors~\cite{yuan2019jointly,khoo2020interpretable, lin2022detect, DBLP:conf/sigir/YangMLG22, ma2020attention, mehta2022tackling}. %propagation tree structure brought insights for \textcolor{blue}{fake news detection,}
%misinformation study, 
%e.g., 
%\citet{ma2018rumor} propose a recursive neural model to bridge the propagation structure with text contents, 
%\citet{yuan2019jointly} encode the semantic and structural information together to detect rumors, and \citet{khoo2020interpretable, DBLP:conf/sigir/YangMLG22} utilize transformer to model user interactions. %In this paper, we propose to retrieve social context information to spot misinformation. %To alleviate tedious eavoid extra work in incorporating external knowledge or distinguishing the meaning of the specific entities, we intend to use a hierarchical model to utilize the social context gathered from social media and detect sentence-level misinformation firstly, and then aggregate the sentence-level results to predict the final article's veracity.
%\textbf{Social-context based methods:}
%Social media allow users to share opinions, conjectures and evidences to emerging events, which provide the crowd of wisdom to spot misinformation. 
The utilization of social context structures has spurred the development of Graph Neural Networks (GNNs) such as Kernel Graph Attention Networks (KGAT)~\cite{liu2020fine} and Graph-aware Co-Attention Networks (GCAN)~\cite{lu2020gcan}, which have demonstrated effectiveness in various fake news-related tasks. 
%\textcolor{blue}{Recent advances in Graph Neural Networks (GNNs) }also help better model the propagation information. \citet{liu2020fine} proposed Kernel Graph Attention Network(KGAT) to detect misinformation, and \citet{lu2020gcan} adopt Graph-aware Co-Attention Networks (GCAN) in misinformation detection task. 
%\textcolor{red}{classify the following papers with their ideas:} 
%\textcolor{blue}{Some other studies turned to social media}, which provide the crowd of wisdom to spot misinformation. \citet{shu2019defend, jin2022towards} developed a sentence-comment co-attention sub-network to select top-k worthy evidences for misinformation detection. \citet{nguyen2020BERTweet} trained a Bert model with enormous tweet comments. 
However, existing approaches~\cite{shu2019defend, jin2022towards, yang2022reinforcement} generally aim to detect article-level fake news, which lack the capability to tell which specific sentences contain misinformation. 
%They cannot be applied to our task because they require sentence-level annotations. %for model training. 
%Therefore, we investigate a weakly supervised framework that retrieves social context information to spot sentence-level misinformation at first and then infer the coarse-level fake news.
%Weakly supervised learning aims to construct predictive models based on incomplete, limited or inaccurate supervision~\cite{keeler1991self}.
%for model training 
%This approach alleviates the burden of data annotation. 

MIL is a weakly supervised approach that infers instance-level labels (e.g., sentence or pixel) when training data is annotated with bag-level labels (e.g., document or image)~\cite{dietterich1997solving}. Several MIL variants have been developed based on threshold-based MIL assumption~\cite{foulds2010review} and weighted collective MIL assumption~\cite{pappas2017explicit}, successfully applied in various downstream tasks such as recommendation systems~\cite{lin2020outfitnet} sentiment analysis~\cite{angelidis-lapata-2018-multiple}, keywords extraction~\cite{wang2016multiple}, community question answering~\cite{chen2017user}, and more recently joint detection of stances and rumors~\cite{DBLP:conf/sigir/YangMLG22}. We adopt the weighted collective MIL assumption~\cite{pappas2017explicit} to incorporate a weight function over the sentence space to calculate the article veracity probability. This assumption allows us to achieve a more robust prediction, as it avoids bias introduced by less important instances. 
%In our task, spotting sentence-level misinformation under the supervision of article-level label is basically a form of weak supervision, which origins from ~\citet{craven1999constructing}, \textcolor{blue}{and efforts to address the issue of massive unlabelled data by utilizing the label signals of other part of data.}
%Following that, weak supervision methods have been adopted in many fields such as computer vision~\cite{li2017learning,patrini2017making,stewart2017label} and natural language process~\cite{natarajan2013learning,bach2017snorkel}. After that, weak labels were used as constraints to regularize classifiers~\cite{hendrycks2018using}, and \citet{zhang2019learning} learned a classifier with weak supervision to identify buggy codes. 
%Nowadays, MIL has become a form of weakly supervised learning method, 
%which is originally defined in a machine learning problem ~\cite{dietterich1997solving}, 
%where instances are arranged in bags and a label accompanied by the entire bag is provided.
%that mainly efforts for classifying binary individual instances (e.g., sentences) in a bag (e.g., article), and then deduce the bag-level prediction by aggregating the instance-level prediction while using only bag-level annotations.
%
%\citet{dietterich1997solving} firstly define Multiple Instance Learning (MIL) as a new machine learning problem, where instances are arranged in bags and a label accompanied by the entire bag is provided. And in this paper we mainly focus on the scenario of Natural Language Processing (NLP).

\section{Problem Definition}
We define a fake news dataset as a set of news articles $\{\mathcal{A}\}$, where each article consists of a set of $n$ sentences $\mathcal{A}=\{s_i\}_{i=1}^n$ and $s_i$ is the $i$-th sentence. For each article, we assume there is a set of $m$ social conversation trees relevant to it denoted as $\mathcal{T}=\{t_j\}_{j=1}^m$, where $t_j$ is the $j$-th conversation tree containing posts (i.e., nodes) and message propagation paths (i.e., edges) which can provide the social context information for $\mathcal{A}$. %We assume that the trees are undirected. 
Our task is to predict the veracity of information at both sentence level and article level in a unified model:
\begin{itemize}[leftmargin=*]
\item
\textbf{Sentence-level Veracity Prediction} aims to determine whether each $s_i \in \mathcal{A}$ is a misinforming sentence or not given its relevant social context information $\mathcal{T}$. That is to learn a function $f(\mathcal{A}): s_1,s_2,\dots,s_n \to p_1,p_2,\cdots p_n$, where $p_i$ is the sentence-level prediction probability as to whether $s_i$ is misinforming or not. 
\item
\textbf{Article-level Veracity Prediction} aims to classify the veracity of the article $\mathcal{A}$ on top of the sentence-level misinformation detection. That is to learn a function $g(\mathcal{A}): p_1,p_2,\cdots p_n \to \hat{y}$, where $\hat{y}$ denotes the prediction as to whether $\mathcal{A}$ is fake or true. Note that we have only article-level ground truth for model training.
\end{itemize}

\section{WSDMS: Our MIL-based Model}

\begin{figure*}[t!]
\centering
\vspace{-0.3cm}
\includegraphics[width=6in]{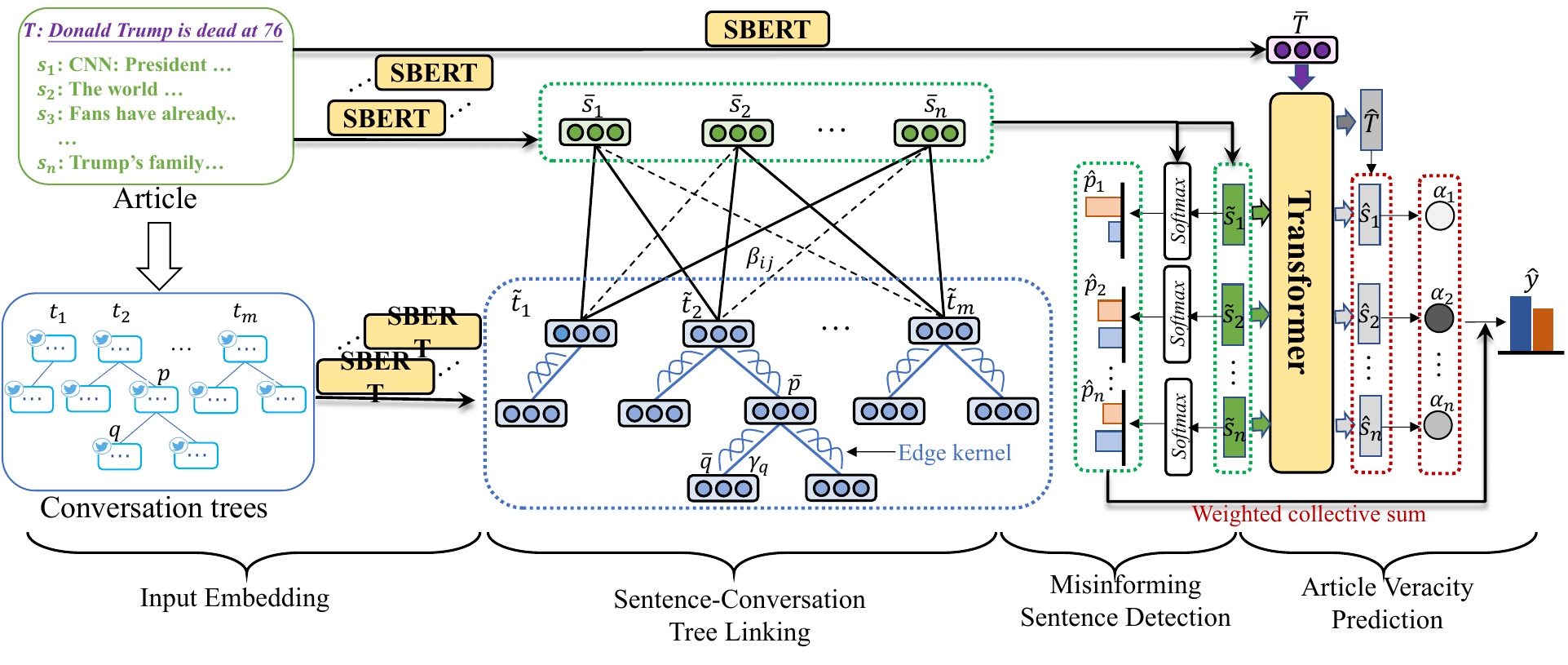}
\caption{The architecture of our WSDMS model. $\bar{t}_i$ denotes the representation of tree $t_i$ after kernel-based interaction of post information among tree nodes.}
\label{fig:model}
\vspace{-0.3cm}
\end{figure*}

%Fake news articles often manipulate factual content by distorting certain parts, creating a blend of truth and falsehood. 
Detecting more nuanced instances of misinformation at the sentence level solely based on article content is challenging~\cite{feng2012syntactic}. Previous studies have demonstrated that social media posts contain valuable opinions, conjectures, and evidence that can be leveraged to debunk claim-level misinformation, such as rumors~\cite{ma2017detect,ma2018rumor,wu2019misinformation}, where claims, typically presented as short sentences, share similar characteristics with sentences in news articles. We hypothesize that the detection of misinforming sentences can be done by incorporating relevant information from social context associated with the article. We try to establish connections between social conversations and specific news sentences in the article, enabling the contextualization of social wisdom to enrich the representation of sentences, in order to better capture the veracity of sentences. 

The architecture of our MIL-based weakly supervised model WSDMS is illustrated in Figure~\ref{fig:model}. WSDMS consists of four closely coupled components: input embedding, sentence and conversation tree linking, misinforming sentence detection, and article veracity prediction. We describe them with detail in this section.

\subsection{Input Embeddings}
 
We represent the word sequence of each news sentence and social post using SBERT~\cite{reimers2019sentence} which maps the sequence into a fixed-size vector. Let a sequence $S=w_1w_2\cdots w_{|S|}$ consist of $|S|$ tokens, where $S$ could optionally denote a news title, a news sentence, or a post in conversation tree. Then, the SBERT embedding of $S$ can be represented by %si={wi1wi2⋯wi|si|}s_i=\{w_{i_1} w_{i_2} \cdots w_{i_|s_i|}\}, ti={wi1wi2⋯wi|ti|}t_i=\{w_{i_1} w_{i_2} \cdots w_{i_|t_i|}\}, where wi,jw_{i,j} is a token which comes from the sentence/post. We then map each wi,jw_{i,j} into a fixed-size hidden vector using BERT~\cite{devlin2019bert} to get the original sentence-level and post-level vector for each sentence or each post. This yields:
%{\setlength{\abovedisplayskip}{0.1cm}
%\setlength{\belowdisplayskip}{0.1cm}
%\begin{equation}\label{equ:PostEncoding}
%\bar{S}=[h_0, h_1, \cdots, h_n]=\text{BERT}(w_1, \cdots, w_n)
%\end{equation}
$\bar{S}=\text{SBERT}(w_1, \cdots, w_{|S|})$.
%where hih_i represents the embedding for the ii-th token in the sequence, and h0h_0 corresponds to the ``[CLS]'' token in BERT pre-trained model w.r.t. the initialized sentence-level embedding for the sequence SS. 
In the rest of the paper, given an article $\mathcal{A}$, we will use additional notations $T$ to denote the news title, $p$ and $q$ to denote posts in a conversation tree. And then $\bar{T}$, $\bar{s}_i$, $\bar{p}$ and $\bar{q}$ will denote the respective SBERT embeddings of $T$, $s_i$, $p$ and $q$.

\subsection{Linking Sentences to Conversation Trees}\label{sec:linking}

% Since misinformation can be blended to seem true by distorting partial contents, spotting sentence-level misinformation becomes a useful and precise step before the coarse-level article verification. 
%Misinformation can be well-edited by distorting parts of factual contents, so it is difficult to spot sentence-level misinformation relying on the article itself~\cite{feng2012syntactic}. Previous studies reveal that social media posts contain opinions, conjectures and evidences to spot misinformation~\cite{wu2019misinformation,ma2018rumor}. Although they focus on coarse-level claim verification, the claim is generally presented as a short sentence that shares similar definitions as our tasks. Therefore, we hypothesize that sentence-level misinformation can be spotted based on the social context information.  %\textcolor{blue}{However, they mainly focus on coarse-level detection on a natural-language input claim, and the claim is generally a relatively short sentence~\cite{popat2018declare}. We can spot sentence-level misinformation by combining the context of the entire article and the information provided in social media posts.}\textcolor{red}{Although they focus on determine whether a claim contains misinformation, why you think it is reasonable to spot misinfo from an article? also refer to the comments in review 1}. 

To mine the discernible relationship between sentences and social posts trees, %in a unified structure after getting the sentence and post encoding. 
we want to design a sentence-tree linking mechanism between the sentence set $\{s_i\}_{i=1}^n$ and post tree set $\{t_j\}_{j=1}^m$, both associated with $\mathcal{A}$. There are clearly different designs to create links across the elements between them, such as 1) using a fully connected graph that links any $s_i$ and $t_j$ regardless of their similarity, followed by a model to fix the closeness of each connection; 2) creating a link according to the similarity between $s_i$ and $t_j$ based on a threshold. %And there are also different ways to compute the sentence-tree similarity, such as only considering the sentence and the root post following~\citet{yang2019unsupervised} or considering the similarity with all posts in the tree. 
Our preliminary experiments indicate that the different designs of interaction indeed influence the performance. Given that the number of sentences and trees associated with articles varies significantly, we opt for the threshold-based approach to avoid the overhead of computing on a fully connected graph.
%suffers from the loss of structural information and . noise, whose propagation pattern may indicate vital misinformation sensitivity~\cite{Yang2022AWS}. \textcolor{red}{show the primary issue of these method towards content matching, that is, why such method fail to match. "simple" is not a problem actually. As for the said "lose information", not clear the reason why the structural info is important.} \textcolor{red}{rewrite the following part:}
%Thus, we adopt the a method to alleviate the propagation information loss: 4) Centralize the information of the post propagation tree to a whole representation, and then connect sentences and post tree root nodes with the retrieval method. %Since the first two methods are simple and may lose some post-structure information,
We begin with modeling posts interaction in each tree to learn its representation before linking the sentences and trees. %resort to \textit{Aggregating node-level flowing information into a centralized tree representation} and then \textit{Link sentences and centralized tree representation} in a unified manner, where the edges between them are undirected.

%\textcolor{red}{what is the relationship of the information passing with our matching problem?}. %Whereas they use a fully connected graph, which is different from our framework, as a result, we resort to \textit{propagating node-level information to get post thread tree representation} and then \textit{link sentences and post thread trees} in a unified manner, and the edges between them are undirected.

%\textcolor{red}{why model propagation among node? why and how it is useful to the linking process? The motivations to choose kernelGAT is required here.}
%\textbf{Aggregating node-level flowing information into a centralized tree representation.} 

\textbf{Post Interaction Embedding:} 
To represent a tree accurately, we use a generic kernel-based graph model KernelGAT~\cite{liu2020fine} to measure the importance of each post in a tree by modeling the interactions between each post and its neighboring posts. 

We first construct a translation matrix $\mathcal{M}$ to represent the similarity of each post with its neighbors, where each $M_{pq} \in \mathcal{M}$ is the cosine similarity between post $p$ and $q$: 
\begin{equation}\label{equ:TranslationMartix}
% \[ 
M_{pq} = \begin{cases}
  \frac{\bar{p}\cdot \bar{q}}{|\bar{p}||\bar{q}|} & \mbox{ if $q \in \mathcal{N}(p)$}\\
  0 & \mbox{ otherwise}
  \end{cases}
% \]
\end{equation}
where $\mathcal{N}(p)$ is the set of neighboring nodes of $p$.

We then define a kernel function 
%\textcolor{blue}{KernelGAT~\cite{liu2020fine} leveraged node matching kernels to measure the importance of evidence node, and modeled finer grained information propagation along graph. To better model fine-grained information flowing and interacting between nodes with different importance and obtain social information representation, we extended KernelGAT to aggregate node-level information in our post tree.} Here we use
$\Vec{\mathcal{G}}(M_p)$ to represent the features considering the interactions between $p$ and its neighbors based on $K$ Gaussian kernels~\cite{keerthi2003asymptotic}, and this yields:
\begin{equation}\label{equ:MatchingKernel}
    \Vec{\mathcal{G}}(M_p) = \left\{\mathcal{G}_1(M_p), \cdots, \mathcal{G}_K(M_p)\right\}
\end{equation}
where
\begin{displaymath}
    \mathcal{G}_k(M_p) = \log\sum_{q \in \mathcal{N}(p)} \exp\left(-\frac{(M_{pq} - \mu_k)^2}{2\sigma_k^2}\right) 
\end{displaymath}
%\sout{where M_{qp} \in \mathbb{R}^{d}M_{qp} \in \mathbb{R}^{d} denotes the token-level similarity between t_qt_q and t_pt_p with \mu_k\mu_k and \sigma_k\sigma_k as parameters in the k^{th}k^{th} kernel to capture the information interactions at different levels~\cite{xiong2017end},}
%\textcolor{blue}{where \vec{K}(M_{q})\vec{K}(M_{q}) utilizes KK kernels to the qq-th row of translation matrix, summarizing it into a KK-dimensional feature vector,} 
and $\mu_k$ and $\sigma_k$ are parameters in the $k$-th kernel to capture the node interactions at different levels~\cite{xiong2017end}. Note that if $\sigma_k \to \infty$, the kernel function degenerates to the mean pooling. %, and \mu_k=1, \sigma_k=0\mu_k=1, \sigma_k=0 \sout{turns into exact matches, }\textcolor{blue}{is equal to TF value.}\textcolor{red}{what is exact matches?}
%\textcolor{blue}{where KK represents the number of matching kernels, and here we set this hyperparameter as 10, each element in matrix M_{qp}M_{qp} is the token embedding similarity between t_qt_q and t_pt_p. And \mu_k\mu_k and \sigma_k\sigma_k are the parameters (mean and width) in k-thk-th kernel to capture the information exchange between tokens at different levels~\cite{xiong2017end}. When \sigma_k \to \infty\sigma_k \to \infty, the kernel function degenerates to the mean pooling, and \mu_k=1, \sigma_k=0\mu_k=1, \sigma_k=0 turns into exact matches.}
%equivalent to the TF value from sparse models.
%the more token pairs with similarities closer to its mean \mu_k\mu_k, the higher its value,
%

Then, we update the representation $\tilde{p}$ of each post $p$ by considering all its neighbors with their identified importance, which is given as: %into consideration: % use u_{q}^{p}u_{q}^{p} to represent post node t_qt_q's fine-grained representation according to the neighbor post node t_pt_p's influence, which can be deduced by:
\begin{equation}\label{equ:Fine-grainedPostEncoding}
\begin{split}
    & \gamma_q = softmax\left(W_1\left(\Vec{\mathcal{G}}(M_p)\right) + b_1\right)[q] \\
    & \tilde{p} = \sum_{q \in \mathcal{N}(p)} \gamma_q \cdot \tilde{q}
    %& u_{q}^{p} = \sum_{i=1}^{\mathcal{S}(q)} \alpha_{i} \cdot h_{q} \\
\end{split}
\end{equation}
where $\gamma_q$ is a scalar representing the post-level attention coefficient between $p$ and its neighbor $q$, $W_1$ and $b_1$ are trainable parameters used to transform $K$ kernels into a vector of all nodes in the tree, $[q]$ takes the value corresponding to post $q$, and $\tilde{p}$ and $\tilde{q}$ are initialized respectively with the BERT-based post embeddings $\bar{p}$ and $\bar{q}$.  

\textbf{Link Sentences and Trees.} With the obtained interaction-enhanced post representations, we use a mean pooling function to represent a conversation tree $t_j$, i.e., $\tilde{t}_j = mean(\sum_p \tilde{p})$ for all $p \in t_j$. 
For each pair of sentences and tree $(s_i,t_j)$ associated with an article, we then create a link between them if the cosine similarity of $\bar{s}_i$ and $\tilde{t}_j$ is above a global threshold $\tau$, where $\tau$ is determined according to the global range of similarity scores between sentences and trees by mapping $\tau$ to the median of the range of scores. %pre-defined threshold. 
We fix this setting empirically.

\subsection{Detecting Misinforming Sentences}

To spot misinforming sentences based on the graph with the sentence-tree links, we propose a graph attention model to detect whether a sentence $s_i$ contains misinformation. Each sentence can be linked to multiple conversation trees and vice versa. In Figure~\ref{fig:intro}, for example, 
%\textcolor{blue}{Our idea aims to spot sentence-level misinformation by incorporating the social context with news sentences. So, we propose a sentence-level graph attention model to determine whether a sentence S_iS_i contains misinformation, after we have built the linking graph between article sentences and social post. Although there are many post trees that focus on the same sentence, the amount of information and importance they can bring vary. As Figure~ shows, 
two trees $t_1$ and $t_3$ are linked to $s_3$, where $t_1$ provides more specific evidence (e.g., the right reward amount and the number of experimental days) indicating that $s_3$ is misinforming, while $t_3$ just implies the sentence is not credible without providing specific clues. Hence, we design an attention mechanism to update the representation of each sentence by considering the importance of all its corresponding trees.%\textcolor{red}{what is the said "coefficient"? what it mean for the task of verify sentence? You should clarify why you calculate the "coefficient". and briefly show the idea of the following equation instead of a simply "inspired by XXX". This is because the following equation seems complex, people may not fully understand.} 

More specifically, let $\mathcal{T}_{i}$ denote the set of trees linked to $s_i$. We aggregate the representation of corresponding trees according to their attention weights, and then update the sentence representation. This is achieved by:  
%\textcolor{blue}{More specifically, we aggregate the representation of post trees according to their importance, and then concatenate the aggregated representation with the initial sentence embedding to form a new sentence representation \Tilde{s}_i\Tilde{s}_i with social context. let \mathcal{T}_{i}\mathcal{T}_{i} denotes the set of post tree nodes connected with sentence S_iS_i,This yields:}
\begin{equation}\label{equ:UpdatedSentenceRep}
\begin{split}
    & \beta_{i,j} = \frac{\exp(\tilde{t}_j \cdot {\bar{s}_i})}{\sum_{t'_j \in \mathcal{T}_i}\exp(\tilde{t'}_j \cdot {\bar{s}_i})} \\
    & \tilde{s}_i = \left(\sum_{{t_j \in \mathcal{T}_i}} \beta_{i,j} \cdot \tilde{t}_{j}\right) \oplus \bar{s}_i
    % & \beta_{j} = Softmax_j( W_1(RELU(W_0(s_i \oplus \Tilde{h}_j))) )
\end{split}
\end{equation}
where $\tilde{s}_i$ denotes the socially contextualized representation of $s_i$, $\beta_{i,j}$ is the importance of $t_j\in \mathcal{T}_i$ with respect to $s_i$, and $\oplus$ denotes concatenation operation. 

We then use a fully-connected softmax layer to predict the probability of $s_i$ containing misinformation based on its BERT-based embedding $\bar{s}_i$ and socially contextualized embedding $\tilde{s}_i$:
\begin{equation}
    \hat{p}_{i} = softmax(W_2\tilde{s}_i + W_3 \bar{s}_i + b_2)
\end{equation}
where $W_2$, $W_3$ and $b_2$ are trainable parameters and $\hat{p}_i$ is the class probability distribution of $s_i$ provided that the bag-level class labels are fake and real, based on the MIL~\cite{foulds2010review,angelidis-lapata-2018-multiple}.

\subsection{Inferring Article Veracity}

We can simply predict an article as fake if there is at least one misinforming sentence is detected, which conforms to the original threshold-based MIL assumption. However, the assumption is overly strong because there can be inaccuracies in sentence-level prediction. % such as \textcolor{blue}{the prediction score of each sentence is very low (indicating potential misinformation), but the veracity of the whole article is actually true.}\textcolor{red}{show an example to explain the bias}. 
Based on the weighted collective MIL assumptions~\cite{foulds2010review}, we design a context-based attention mechanism to bridge the inconsistency between sentence- and article-level predictions. 

Specifically, we first learn a global representation for the article utilizing a pre-trained transformer~\cite{grail2021globalizing}:
\begin{equation}\label{equ:interactive sentence-awareArticleRepresentation}
    [\hat{T}, \hat{s}_1, \cdots, \hat{s}_n] = Trans\left([\bar{T}, \tilde{s}_{1}, \cdots, \tilde{s}_{n}]\right)
\end{equation}
where $\bar{T}$ is the initial SBERT embedding of the article title.
We then adopt an attention mechanism to measure the importance of sentences w.r.t the article veracity prediction, which yields: %which can be achieved by:
\begin{equation}\label{equ:interactive sentence-awareAttention}
\begin{split}
    & \alpha_i = \frac{\exp(\hat{s}_i \cdot \hat{T})}{\sum_{i=1}^n \exp(\hat{s}_i \cdot \hat{T})} \\
    & \hat{y} = \sum_{i=1}^n \alpha_i \cdot \hat{p}_i
\end{split}
\end{equation}
where $\alpha_i$ denotes the attention weight of $\hat{s}_i$ relative to the title representation $\hat{T}$, and $\hat{y}$ is the class probability distribution of $\mathcal{A}$ being fake or real. 

\subsection{Model Training} \label{Sec:LossFunction}
Intuitively, the more similar two sentences are, the more similar their corresponding predictions should be. We define the following loss function considering pairwise consistency between sentence representation and prediction, with only article-level ground truth: 
\begin{equation}\label{equ:loss}
    \mathcal{L}(\mathcal{A}) =  \lambda \cdot \mathcal{C}(\mathcal{A})+(1-\lambda) \cdot ||y_{\mathcal{A}}-\hat{y}_{\mathcal{A}}||_2^2
\end{equation}
where 
% % \begin{equation}\label{equ:PairWiseMeasure}
\begin{displaymath}
   \mathcal{C}(\mathcal{A}) = \sum_{i=1}^n \sum_{j=1}^n \exp \left(-||\hat{s}_i-\hat{s}_j||_2^2 \cdot ||\hat{p}_i- \hat{p}_j||_2^2 \right)
\end{displaymath}
% \end{equation}
Here $\mathcal{C}(.) \in [0,1]$ is the function measuring the consistency between pairwise sentence similarity (i.e., $\hat{s}_i$ and $\hat{s}_j$) and the prediction (i.e., $\hat{p}_i$ and $\hat{p}_j$), $y_{\mathcal{A}}$ and $\hat{y}_\mathcal{A}$ denote respectively the ground-truth and predicted class probability distributions of $\mathcal{A}$, $||.||_2^2$ is an efficient kernel based on the L2 norm~\cite{luo2016regression} as a non-negative penalty function, and $\lambda$ is the trade-off coefficient.
%So $\sigma(\hat{p}_{i}, \hat{p}_{j})= (\hat{p}_{i}-\hat{p}_{j})^2$ represents non-negative penalty on the difference between predictions from sentence $S_i$ and $S_j$, $\sigma(y_n, \hat{y}_n)=(y_n-\hat{y}_n)^2$ represents non-negative penalty on the difference between prediction and ground-truth of article $n$, and $\lambda$ is the trade-off coefficient. 

\section{Experiments and Results}

\subsection{Datasets and Setup}
We employ two public real-world datasets PolitiFact and GossipCop~\cite{shu2020fakenewsnet} respectively related to politics and entertainment fake news, where relevant social conversations are collected from Twitter. We also construct an open-domain fake news dataset BuzzNews by extending  BuzzFeed~\cite{tandoc2018five}, for which we gather social conversations of the articles via Twitter API\footnote{\url{https://developer.twitter.com/en/docs}}. 

We recruit three annotators to label misinforming sentences of the articles in the test sets of the three datasets. We train the annotators by providing them with a unified set of annotation rules referring to the detailed guide from several fact-checking websites such as snopes.com and politifact.com, where specific rationales on how each claim was judged are provided. Then, we take a majority vote for determining the label of each sentence, and the inter-annotator agreement is 0.793.
%mark the result by cross-checking to avoid the interference of individual ideology and position\footnote{The average time of labeling one article is approximately 2 mins.}.} 
Table~\ref{tab:dataset} shows the statistics of these three datasets.

\begin{table}[t!]
\small
%\vspace{-0.5cm}
%\setlength{\abovecaptionskip}{-0.2cm}
%\setlength{\belowcaptionskip}{-0.8cm}
  \centering
  \resizebox{0.48\textwidth}{!}{
   \vspace{-0.2cm}
    \begin{tabular}{lrrrr}
    \toprule
          & \textbf{Stat.} & \textbf{PolitiFact} & \textbf{GossipCop} & \textbf{BuzzNews} \\
    \midrule
    \multirow{2}{*}{\textbf{Train}} & \# True & 624   & 16,658 & 301 \\
          & \# Fake & 432   & 5,255 & 105 \\
    \midrule
    \multirow{2}{*}{\textbf{Test}} & \# True & 140   & 160   & 50 \\
          & \# Fake & 70    & 80    & 25 \\
    \midrule
    \textbf{Total} & --  & 1,270 & 22,153 & 481 \\
    \midrule
          % & \#sent & 38,191 & 607,590 & 12,871 \\
          % & \#trees & - & - & - \\
          % & \#tweets & 400,832 & 1,291,236 & 163,432 \\
    \multicolumn{2}{r}{\# avg. sent/art}  & 30  & 27  & 27 \\
    \multicolumn{2}{r}{\# avg. trees/art}  & 13  & 16  & 9 \\
    \multicolumn{2}{r}{\# avg. posts/tree}  & 316  & 58  & 340 \\
    \bottomrule
    \end{tabular} }%
%    \vspace{-0.2cm}
  \caption{Statistics of the datasets used.} 
  \vspace{-0.3cm}
  \label{tab:dataset}%
  \vspace{-0.3cm}
\end{table}%

%We use precision, recall, F1, and accuracy to as evaluation metrics. 
We use precision (Pre), recall (Rec), F1, and accuracy (Acc) as evaluation metrics. 
All the baselines and our methods are implemented with PyTorch~\cite{paszke2019pytorch} (see Appendix~\ref{sec:appendix2} for implementation details).

% Our model parameters are updated by back-propagation~\cite{collobert2011natural} with Adam~\cite{DBLP:journals/corr/KingmaB14} optimizer. We set the maximum epoch to 100, dimension of embeddings to 512 for sentence and posts, and empirically initialize the learning rate as 0.001, and the hyperparameter λ\lambda is set to 0.5 which is validated on a small hold-out dataset. 
% The training process ends when the loss value converges or the maximum epoch number is met. 
% %\footnote{The maximum epoch is set as 150 in our experiment.}.
% % For each dataset, we manually label some fine-grained level labels for testing.

\subsection{Article-level Fake News Detection} \label{sec:Coarse-levelResult}

\begin{table*}[t!]
\setlength{\belowcaptionskip}{-0.5cm}
  \centering
  \small
  \resizebox{1.0\textwidth}{!}{
  \vspace{-0.2cm}
    \begin{tabular}{l|cccc|cccc|cccc}
    \toprule
    \textbf{Dataset} & \multicolumn{4}{c|}{\textbf{PolitiFact}} & \multicolumn{4}{c|}{\textbf{GossipCop}} & \multicolumn{4}{|c}{\textbf{BuzzNews}} \\
    \midrule
    \textbf{Method} & \textbf{Pre} & \textbf{Rec} & \textbf{F1} & \textbf{Acc} & \textbf{Pre} & \textbf{Rec} & \textbf{F1} & \textbf{Acc} & \textbf{Pre} & \textbf{Rec} & \textbf{F1} & \textbf{Acc} \\
    \midrule
    DeClarE & 0.714  & 0.746 & 0.730  & 0.789 & 0.706 & 0.741 & 0.723 & 0.762  & 0.705  & 0.743  & 0.724  & 0.754  \\
    HAN   & 0.752 & 0.779 & 0.765 & 0.803 & 0.718 & 0.739 & 0.728 & 0.789  & 0.727  & 0.768  & 0.747  & 0.762  \\
    dEFEND & 0.900  & 0.926 & 0.913 & 0.886 & 0.729 & 0.785 & 0.756 & 0.808  & 0.731  & 0.792  & 0.760  & 0.810  \\
    BerTweet & 0.844 & 0.903 & 0.873 & 0.878 & 0.851 & 0.862 & 0.857 & 0.848  & 0.831  & 0.840  & 0.835  & 0.811  \\
    \midrule
%    H-GCN & 0.720  & 0.735 & 0.727 & 0.800  & 0.708 & 0.720  & 0.714 & 0.784  & 0.699  & 0.714  & 0.706  & 0.759  \\
    GCAN  & 0.817 & 0.821 & 0.819 & 0.837 & 0.782 & 0.803 & 0.792 & 0.791  & 0.780  & 0.800  & 0.790  & 0.795  \\    
    Bi-GCN & 0.852 & 0.838 & 0.845 & 0.865 & 0.797 & 0.813 & 0.805 & 0.822  & 0.791  & 0.814  & 0.802  & 0.817  \\
    KAN   & 0.870  & 0.840  & 0.855 & 0.859 & 0.776 & 0.770  & 0.773 & 0.807  & 0.766  & 0.790  & 0.778  & 0.820  \\
    SureFact & 0.913 & 0.939 & 0.924 & 0.887 & 0.859 & 0.872 & 0.865 & 0.847  & 0.841  & 0.856  & 0.848  & 0.829  \\
    \midrule
    \textbf{WSDMS} & 0.921 & \textbf{0.967} & 0.943 & 0.904 & \textbf{0.864} & 0.876 & 0.870  & 0.850  & \textbf{0.850} & \textbf{0.857} & \textbf{0.853}  & \textbf{0.858} \\
%    \textbf{HotCake-SW} & 0.921 & 0.967 & 0.943 & 0.904 & 0.864 & 0.876 & 0.870  & 0.852  & 0.850  & 0.857  & 0.853  & 0.858  \\
    \textbf{WSDMS-FC} & \textbf{0.923} & \textbf{0.967} & \textbf{0.944} & \textbf{0.908} & 0.862 & \textbf{0.879} & \textbf{0.871} & \textbf{0.853} & \textbf{0.850}  & \textbf{0.857}  & \textbf{0.853}  & \textbf{0.858}  \\
    \bottomrule
    \end{tabular}}%
    \vspace{-0.2cm}
   \caption{Article-level fake news detection results.} % \textcolor{red}{p-value < 0.05 under t-test (5 different dataset splits)}.}
  \label{tab:coarselevelresults}%
  % \vspace{-0.1cm}
\end{table*}%

We compare the following models at the article level. Some original settings of baselines might not suit the data in this task, which have to be specifically customized (see Appendix~\ref{sec:appendix1}).
1) \textbf{DeClarE}~\cite{popat2018declare}: An evidence-aware network using news title to attend over words in relevant posts for verifying news claims. %\textcolor{red}{DeClarE originally uses new title and article for fake news detection. Why do you use posts to replace news content as said in the appendix. I believe these are cases that baselines' original setting is not respected.}
% \item \textbf{MT-GRU}~\cite{ma2018detect}: A multi-task learning approach based on GRU for joint detection of rumors and stances by capturing the both shared and task-specific features. Here we treat sentence veracity as stance veracity. 
2) \textbf{HAN}~\cite{ma2019sentence}: A hierarchical attention network using the news title to attend over relevant posts as evidences.
3) \textbf{dEFEND}~\cite{shu2019defend}: A sentence-post co-attention network for fake news detection. 
4) \textbf{BerTweet}~\cite{nguyen2020BERTweet}: A language model pre-trained on 850M tweets, which is applied here for article verification using article and relevant posts.
%\textcolor{red}{how it's applied here? Applied on claim, news article or posts? If on posts, what posts are used? An article can have multiple trees. Is it applied on the linked trees or all tree of the article?}
% \textcolor{red}{A pre-trained language model applied for claim verification.} 
%5) \textbf{H-GCN}~\cite{wei2019modeling}: \textcolor{red}{A hierarchical multi-task learning framework for jointly predicting coarse-level and finer-level misinformation.} %with Graph Convolutional Networks. need sentence labels.
5) \textbf{GCAN}~\cite{lu2020gcan}: A graph-aware co-attention model trained on user profile and post propagation structure without using post content to verify the news given title.
% \textcolor{red}{A graph-aware co-attention model utilizing retweet to verify the source claim.} 
6)  \textbf{Bi-GCN}~\cite{bian2020rumor}: A bi-directional graph convolutional network using news title and propagation structure of posts for verifying the news.
%\textcolor{red}{Less clear than previous version: what are exactly used? Post content or tree structure only? I assume such methods, when applied on your data, didn't consider article content and title?}
% \textcolor{red}{State-of-the-art rumor detection model utilizing Bi-directional GCN over the social context.}
7) \textbf{KAN}~\cite{dun2021kan}: An attention network utilizing entities in article content and entity contexts for fake news detection.
8) \textbf{SureFact}~\cite{yang2022reinforcement}: A reinforcement subgraph reasoning method using the topic connection between article and relevant posts for fake news detection.
9) \textbf{WSDMS}: Our proposed weakly supervised method. %that utilizes interactive sentence-aware attention network to detect coarse-level article veracity.
%11) \textbf{WSDMS-SW}: A variant of our method that treats two sentences as a unit using a sliding window of size 2\footnote{Because two or more true sentences together may contain misinformation by some logistic errors.}. 
10) \textbf{WSDMS-FC}: A variant of our method that fully connects sentences and post trees. 
%\ruichao{The specific information used by these baselines is different, so, we illustrate these baselines and how our data is applied to them exhaustively in the ??????\ref{sec:appendix1}.}\textcolor{red}{Overall, while you list numerous baselines, it's unclear what make them comparable with your method. Feeding them different portions of data will definitely affect their performance, but that may make the comparison unfair as they used different info at input.}
% \end{itemize}
Table~\ref{tab:coarselevelresults} presents the following observations: 
\begin{itemize}[leftmargin=*]
\item In the first group of structured models, dEFEND performs the best. This is because DeClearE and HAN are designed to only use the \textit{external} relevant context of a claim and BerTweet is trained to represent social posts. dEFEND leverages features extracted from both article content and external posts that are complementary. %BerTweet is the second best due to the strong BERT-based text representation.
%DeClarE performs worst because it attends on words in posts but ignores the social context information. %non-structured and structured baselines, this is because tweets are almost short texts, and the context of texts may be lost for the article, resulting in inaccurate calculation of per-article credibility of claim. 
%dEFEND outperforms HAN, indicating that the features extracted from the article and posts are complementary. BerTweet achieves a higher F1 score because it is a pre-trained model with superior text representations. 
\item In the second group of non-structured models, the graph-based models GCAN and Bi-GCN mainly rely on propagation structures of fake news and perform comparably with KAN using entities and their contexts extracted from the social media content, suggesting that social conversations embed a good amount of human wisdom useful for detecting fake news. SureFact performs best among all the baselines because it groups social posts into the topics discovered from article content, suggesting that creating a connection between them at the topic level is helpful.
%\textcolor{red}{SureFact is the best among all the baselines because it groups structured comments according to the topic and filters evidential comments, suggesting that social contexts play a vital role in inferring fake news.} 
%\textcolor{blue}{Although H-GCN utilizes the structure information, it only considers one adjacent node, resulting in worst performance among structured methods. GCAN and Bi-GCN outperforms H-GCN, indicating that we can identify some useful patterns from the propagation structure to indicate fake news. KAN uses external knowledge and performs better than Bi-GCN, indicating that the introduction of external knowledge can supplement additional information to help to detect fake news. SureFact performed best in the method of using structure, because it groupes structured comments according to topic, and then used RL method to select comments for hot topic as evidence, indicating that hot-topic related social information can also help to locate fake news accurately. But one point we need to note is that the performance of dEFEND on PolitiFact is even better than the structured method KAN, due to the external knowledge introduced by KAN may also introduce a lot of noise, and the quality of comments on PolitiFact is higher compared with the other two datasets. This demonstrates that social posts play a vital role in inferring fake news, even surpassing external knowledge, which is consistent with our motivation.} 
\item WSDMS consistently defeats the best baseline SureFact on the three datasets, demonstrating that our explicit and fine-grained linking between sentence and social context is superior, and the sentence-level detection can help article veracity prediction. In addition, WSDMS does not sacrifice its performance compared to WSDMS-FC that uses full connections between sentences and trees, while we find that WSDMS significantly reduces training time from 4.5 to 2 hours. This indicates our sentence-tree linking method is cost-effective.
\end{itemize}

\subsection{Misinforming Sentence Detection}\label{Sec:Sentence-levelResult}
%We compare our sentence-level misinformation detection model with some existing systems, the following state-of-the-art methods. 
%Since our task is newly defined and existing work only aimed at detecting coarse-level claim/article, to make a fair comparison with state-of-the-art baselines, we take each sentence as a claim and implement all the baselines with the same settings as ours. % with PyTorch~\cite{DBLP:journals/corr/KingmaB14} for a fair comparison. 

\begin{table*}[t!]
\setlength{\belowcaptionskip}{-0.3cm}
  \centering
  \small
  \resizebox{1.0\textwidth}{!}{
  \vspace{-0.2cm}
    \begin{tabular}{l|cccc|cccc|cccc}
    \toprule
    \textbf{Dataset} & \multicolumn{4}{c|}{\textbf{PolitiFact}} & \multicolumn{4}{c|}{\textbf{GossipCop}} & \multicolumn{4}{|c}{\textbf{BuzzNews}} \\
    \midrule
    \textbf{Method} & \textbf{Pre} & \textbf{Rec} & \textbf{F1} & \textbf{Acc} & \textbf{Pre} & \textbf{Rec} & \textbf{F1} & \textbf{Acc} & \textbf{Pre} & \textbf{Rec} & \textbf{F1} & \textbf{Acc} \\
    \midrule  
    DeClarE & 0.504  & 0.531  & 0.517  & 0.559  & 0.501  & 0.528  & 0.514  & 0.550  & 0.513  & 0.520  & 0.516  & 0.540  \\
%    MT-GRU & 0.529  & 0.559  & 0.544  & 0.580  & 0.531  & 0.550  & 0.540  & 0.568  & 0.520  & 0.543  & 0.531  & 0.544  \\
    HAN   & 0.531  & 0.559  & 0.545  & 0.565  & 0.510  & 0.529  & 0.519  & 0.561  & 0.518  & 0.537  & 0.527  & 0.562  \\
    dEFEND & 0.539  & 0.586  & 0.562  & 0.605  & 0.534  & 0.581  & 0.557  & 0.600  & 0.538  & 0.570  & 0.554  & 0.580  \\
    BerTweet & 0.542  & 0.630  & 0.583  & 0.619  & 0.539  & 0.619  & 0.576  & 0.602  & 0.542  & 0.610  & 0.574  & 0.599  \\
    \midrule
%    H-GCN & 0.520  & 0.543  & 0.531  & 0.564  & 0.507  & 0.531  & 0.519  & 0.558  & 0.511  & 0.530  & 0.520  & 0.551  \\
    GCAN  & 0.533  & 0.563  & 0.548  & 0.589  & 0.511  & 0.561  & 0.535  & 0.581  & 0.521  & 0.551  & 0.536  & 0.580  \\  
    Bi-GCN & 0.557 & 0.589 & 0.573 & 0.606 & 0.531 & 0.560 & 0.545 & 0.593 & 0.533 & 0.553 & 0.543 & 0.601  \\    
    KAN   & 0.574  & 0.594  & 0.584  & 0.611  & 0.539  & 0.561  & 0.550  & 0.609  & 0.540  & 0.560  & 0.550  & 0.610  \\
%    SureFact & ? & ? & ? & ? & ? & ? & ? & ? & ? & ? & ? & ?  \\ 
    \midrule
    WSDMS & 0.518 & 0.539 & 0.527 & 0.564 & 0.508 & 0.531 & 0.519 & 0.562 & 0.513 & 0.537 & 0.524 & 0.549\\
    \midrule
    \textbf{WSDMS (o)} & 0.637  & 0.676  & 0.655  & 0.644 & 0.629  & \textbf{0.664} & 0.646  & \textbf{0.639}  & 0.609 & 0.587  & 0.598  & \textbf{0.662} \\
%    \textbf{HotCake-SW} & 0.637  & 0.675  & 0.655  & 0.650  & 0.630  & 0.664  & 0.647  & 0.639  & 0.609  & 0.587  & 0.598  & 0.562  \\
    \textbf{WSDMS-FC (o)} & \textbf{0.639} & \textbf{0.679} & \textbf{0.658} & \textbf{0.650} & \textbf{0.633} & \textbf{0.664}  & \textbf{0.648} & \textbf{0.639} & \textbf{0.610}  & \textbf{0.590} & \textbf{0.600} & \textbf{0.662}  \\
    \bottomrule
    \end{tabular}}%
    \vspace{-0.2cm}
   \caption{Minformaing sentence detection results.} % \textcolor{red}{p-value < 0.05 under t-test (5 different dataset splits)}.}
  \label{tab:sentencelevelresults}%
  \vspace{-0.3cm}
\end{table*}% 

For misinforming sentence detection, the baselines are deployed by treating each sentence as a claim and the conversation trees linked to the sentence (see Section~\ref{sec:linking}) as the source of evidence. SureFact is excluded as it cannot classify specific sentences. More details are in Appendix~\ref{sec:appendix1}.
%\textcolor{red}{The setting in the table seems incorrect or inconsistent which may lead unfair comparisons. Need to talk further.}
%We compare the following models:
%1) \textbf{MT-GRU}~\cite{ma2018detect}: A multi-task learning approach based on GRU-RNN model for jointly detecting rumors and stances (we obtain the sentence spotting by transforming stances). 
%(1) \textbf{DeClarE}, \textbf{HAN}, \textbf{dEFEND}, \textbf{BerTweet}, \textbf{GCAN}, \textcolor{red}{\textbf{Bi-GCN}} and \textbf{KAN}: We treat each sentence as a claim, and their related trees as additional shares of evidence, applied for spotting sentence-level misinformation.
% (2) \textbf{DeClarE}: An evidence-aware deep method to debunk false claims. 
% (3) \textbf{dEFEND}~\cite{shu2019defend}: A sentence-comment co-attention network for fake news detection. 
% (4) \textbf{BERTweet}~\cite{nguyen2020BERTweet}: A pre-trained language model to verify source claim. 
% (5) \textbf{GCAN}~\cite{lu2020gcan}: A graph-aware co-attention model utilizing retweet structure to verify the source claim. 
% (6) \textbf{H-GCN}~\cite{wei2019modeling}: A hierarchical multi-task learning framework for jointly predicting coarse level and fine-grained level misinformation with graph convolutional network. 
% (7) \textbf{KAN}~\cite{dun2021kan}: An attention network that uses external knowledge to detect fake news. 
%(3) \textbf{WSDMS} and \textbf{WSDMS-FC}: Our proposed weakly supervised method and two variations by utilizing social posts to detect sentence-level misinformation. 

Since all baselines are supervised methods that need sentence labels for training, we split the three test sets with sentence-level annotation into train and test parts with a 70\%-30\% ratio.
%\textcolor{red}{These baselines are not designed for sentence prediction, and if they're applied here, they need sentence labels. Then why do you need to compare with them? Also, if they're trained on labeled sentences, is your model also trained on these 70\% data, or still on the original training set? If they train on the same 70\%data, why baselines trained with sentence labels perform worse than your model indirectly trained on article labels? Is your model is trained on the original training set, then it's clearly incomparable with baselines. This can be challenged.}. 
Due to the large number of sentences in the original test sets (6,300/6,480/2,480), we end up with three workable sentence-level training and test sets.
%\textcolor{red}{how many sentences are there in the test sets, \# of fake/true sentences?}. 
We then train all models on the same training data. But this intentionally disadvantages our WSDMS since it can only use article labels.  
Therefore, we also present the performance of WSDMS (o) trained on the original training sets without sentence labels, which baselines cannot take advantage of.
Table~\ref{tab:sentencelevelresults} conveys the following findings: %methods in the first group only incorporate content information with sequential structure.
\begin{itemize}[leftmargin=*]
\item Similar to article-level prediction, dEFEND outperforms DeClarE and HAN because it effectively models the sentence and social context correlations via the co-attention mechanism. BERTweet is more advantageous at representing social media posts, demonstrating better performance at the sentence level. %  performs best because it is a pre-trained language model trained on a , introducing a better generalization ability and good performance on all three datasets.}\textcolor{red}{update this sentence according to Sec 5.2}.  
%
%In the second group, all the methods incorporate textural contents with specified propagation structure on social network. 
\item Among the structured models, KAN performs best because it incorporates both content and propagation information and has a co-attention mechanism between sentence and entity contexts extracted from social conversations. This may enhance sentence representation better than Bi-GCN and GCAN that can only utilize propagation-based features. %\textcolor{red}{say something about why they are not effective for sentence-level detection, consistent with your motivation why propose a new model} 

%However, all the baselines are designed for detecting coarse-level claims/articles, which cannot be readily and effectively applied to spot sentence-level misinformation. Thus their performances are limited, which motivates us to propose a new method. 
\item Weakly supervised WSDMS performs better than DeClarE and comparably with HAN, which are fully supervised. This is because WSDMS considers the propagation structure while DeClarE and HAN can only leverage unstructured posts. The overall performance of WSDMS is clearly compromised due to weak supervision. 
%which is below the average level of supervised baselines.
However, when it is trained on the original datasets, WSDMS (o) can enjoy the large volume of article labels to beat all baselines that cannot be weakly supervised. To reach the same level of performance, the baselines may need tremendous sentence annotations which are infeasible to get. %\ruichao{performs best compared with all baselines on all three datasets, this is because our method models content and context information with indicative misinformation characteristic, under the auxiliary of social context.}
Again, it performs comparably well as WSDMS-FC (o), implying that our sentence-tree linking reserves vital information for spotting misinforming sentences efficiently. 

\item WSDMS effectively enhances sentence-level performance by utilizing publicly accessible article-level labels. To achieve comparable performance, baseline systems generally require massive fine-grained sentence-level annotations. Consequently, sentence-level prediction remains a pivotal contribution of our study. %WSDMS demonstrates its capacity to boost sentence-level performance under the supervision of readily accessible article-level labels, while the conventional baseline methods cannot reach. For baselines to get comparable performance, there is an impractical high bar to prepare fine-grained sentence-level annotations. As a result, the aspect of sentence prediction constitutes one of the primary contributions of our study.}

\end{itemize}

\subsection{Ablation Study}
We ablate WSDMS based on the PolitiFact dataset by varying some component(s): 
1) \textbf{w/o $\boldsymbol{\tau}$}: Fully connect sentences and trees by removing $\tau$, i.e., WSDMS-FC.
2) \textbf{w/o NLL}: Replace the loss with an ordinary negative log-likelihood loss function. 
3) \textbf{w/o wc}: Infer article veracity based on the original MIL assumption without weighted collective attention. 
4) \textbf{Title as sent}: Treat the title as a common sentence.
5) \textbf{w/o kernel}: Reduce the kernel-based post interaction embedding to dot-product attention between sentence and conversation trees. 
6) \textbf{w/o tree}: Remove conversation trees. 

\begin{figure}[t!]
\setlength{\belowcaptionskip}{-0.2cm}
% \centering
\includegraphics[width=2.9in]{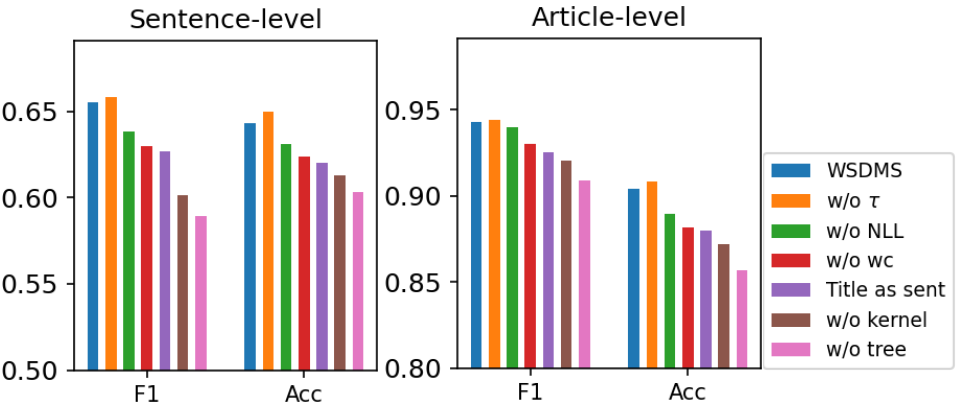}
\caption{Ablation results on PolitiFact dataset.}%\textcolor{red}{update legend names}} 
\label{fig:ablationstudy}
\vspace{-0.2cm}
\end{figure}

Figure~\ref{fig:ablationstudy} shows that most of the ablations make the result worse. \textbf{w/o tree} implies that only using article content is insufficient for the task. \textbf{w/o kernel} supports that embedding post interactions with kernel can help post and tree representation. Experiment in the Appendix~\ref{sec:apendix3} also echoes the advantages of the kernel. \textbf{Title as sent} means that the news title may attract the most attention from the trees, which can hurt the representation of other sentences, and should be specially treated. \textbf{w/o wc} indicates adopting weighted collective MIL is better. \textbf{w/o NLL} confirms that our designed loss is necessary and effective. Only \textbf{w/o {$\boldsymbol{\tau}$}} is marginally better due to fully connected sentences and trees, which is however more costly and less efficient.

\subsection{Case Study}
\begin{figure}[t!]
\setlength{\belowcaptionskip}{-0.2cm}
% \centering
\includegraphics[width=3.1in]{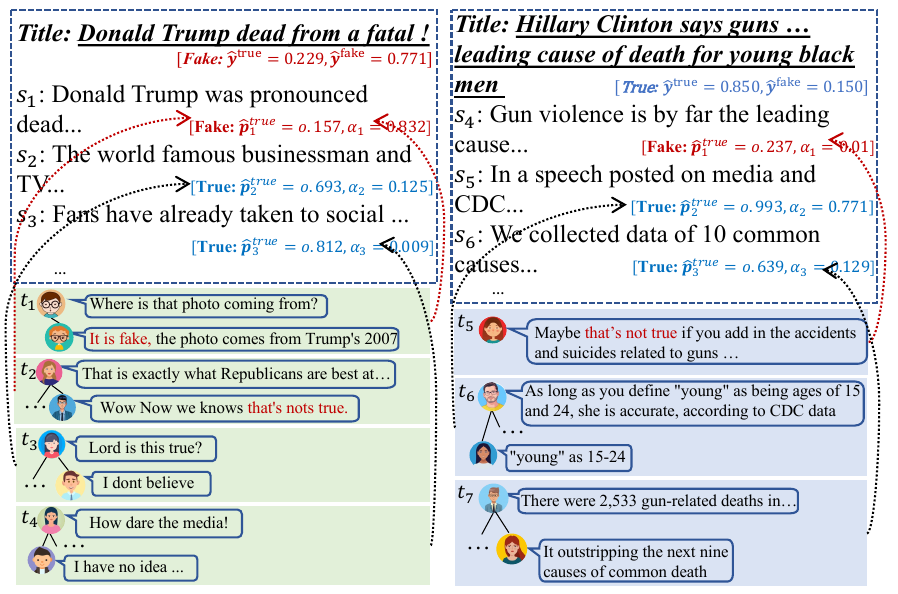}
\caption{A case study illustrating the prediction.} 
\label{fig:CaseStudy}
\vspace{-0.2cm}
\end{figure}

To gain a deeper insight, we visualize two news articles checked by PolitiFact in Figure~\ref{fig:CaseStudy} which are predicted as fake (left) and true (right) correctly by WSDMS.
% a fake news article debunked by PolitiFact and is also predicted as fake correctly by WSDMS, together with 
The spotted misinforming and true sentences are also shown. %we observe that and sample a Misinformation sentence and fact sentence separately to show the sentence-level misinformation and coarse-level article verification results. Table~ shows the connection details between the article and the relevant posts propagation tree. 
We observe that  
1) WSDMS can associate a sentence with multiple trees using attention weights (arrow lines indicate high-weight trees) to help determine its veracity. 2) The posts in the conversations provide useful clues for indicating how credible each sentence is by aggregating collective opinions of users in the trees; 
3) The article-level veracity is not determined simply by whether there is a misinforming sentence detected, because the prediction might be inaccurate. For example, if $s_4$ is incorrectly predicted as fake, the article will also be determined as fake under the standard MIL. Our approach increases the chance of correcting such an error by giving higher attention weights to other sentences, which may indicate that the article is overall more likely to be true. Thus, the attention weights of sentences can collectively aggregate sentence-level predictions to improve the final prediction.
%\textcolor{red}{But the naive MIL also works well in this case. How this example can explain collective assumption is better than naive?} %and the key to identifying coarse-level article veracity is to distinguish misinformation that may be blended to seem true by distorting partial contents, and sentence-level misinformation can be spotted via weakly supervised learning with only the coarse-level article veracity label.

\subsection{User Study Experiment} \label{sec:appendix4}

% misinformation result in fake news can be described as a phenomenon observed in results.
% Table generated by Excel2LaTeX from sheet 'user study'
\begin{table}[htbp]
  \centering
  \small
 \resizebox{0.48\textwidth}{!}{
%  \vspace{-0.2cm}
    \begin{tabular}{lcccc}
    \toprule
    & \textbf{F1}    & \textbf{Acc}   & \textbf{Confidence} & \textbf{Avg. Time/news} \\
    \midrule
    \textbf{Baseline} & 0.784 & 0.795 & 2.017 & 10 sec \\
    \textbf{WSDMS} & 0.958 & 0.989 & 3.206 & 3 sec \\
    \bottomrule
    \end{tabular}%
    }
   \vspace{-0.2cm}
  \caption{User study results on model outputs quality.}
  \label{tab:UserStudy}
  \vspace{-0.2cm}
\end{table}

We conduct a user study to evaluate the quality of the model output. We sample 120 articles from PolitiFact and present them in two forms: Baseline (article, posts) and WSDMS (article, misinforming sentences, trees).
%in a tuple including article, spotted sentences, and related trees linked with each sentence. 
We then ask 6 users to label the articles and give their confidence in a 5-point Likert Scale~\cite{joshi2015likert}, and each person is given only one form to avoid cross influence. 

Table~\ref{tab:UserStudy} shows that 1) users determined the article-level veracity more accurately with WSDMS; 2) users spent 70\% less time identifying fake news; and 3) users showed higher confidence with the results of WSDMS, suggesting they tend to be more sure about their decisions when specific misinforming sentences and relevant evidence are provided.

\section{Conclusion and Future Work}
We propose a MIL-based model called WSDMS to debunk fake news in a finer-grained manner via weakly supervised detection of misinforming sentences with only article veracity labels for model training. WSDMS uses the attention mechanism to associate news sentences with their relevant social news conversations to identify misinforming sentences and determine the article's veracity by aggregating sentence-level predictions. %a hierarchical embedding model to link the article and related posts at first, then a sentence-level graph attention model to spot , finally an interactive sentence-aware attention model to detect coarse-level article veracity. 
WSDMS outperforms a set of strong baselines at the article level and sentence level on three datasets. 

In the future, we will incorporate more inter-sentence features, such as discourse relations, to detect composition-level misinformation.
%Finally, we collect a new misinformation dataset with both sentence-level and coarse-level annotations. % for future studies in fine-grained misinformation detection. %In the future, we want to incorporate more external knowledge such as user profiles or knowledge graphs to help combat misinformation and fake news with coarse-to-fine weakly supervised learning.

\section*{Limitations}
Fake news is one type of misinformation, which also includes disinformation, rumors, and propaganda. WSDMS can be well-generalized to detect these various forms of misinformation. %can detect misinformation and disinformation carrying an additional harmful intent at the sentence level. And in this paper, we clarify that disinformation is a subset of misinformation that our approach can generally deal with.}
Whereas, we simplify some techniques in this paper. For example, the representation of conversation trees can be learned by considering the direction of message propagation and combining top-down and bottom-up propagation trees. 
%Moreover, the parameter settings in the model can also be adjusted again, such as lr, epoch, etc. 
In addition, it cannot deal with more complex situations, where multiple true sentences combined constitute some kind of logical falsehoods or inconsistencies. This can be strengthened by considering sentence-level relations such as discourse information in the model. Despite this limitation, WSDMS encounters no such situation in the three datasets used according to our observation. Nevertheless, this suggests that the existing fake news datasets and detection models lack consideration of discourse-level fakes or logically inconsistent compositions, which are presumably not uncommon in real-world fake news. Lastly, we only use social context data collected from Twitter, which might have platform bias. To mitigate the issue, we can introduce additional data from different social media platforms, such as BuzzFace~\cite{santia2018buzzface} from Facebook.

\section*{Acknowledgements}
This work was partially supported by Hong Kong RGC ECS (Ref. 22200722), National Natural Science Foundation of China Young Scientists Fund(No. 62206233) and HKBU One-off Tier 2 Start-up Grant (Ref. RCOFSGT2/20-21/SCI/004).

\section*{Ethics Statement}
%Scientific work published at EMNLP 2023 must comply with the \href{https://www.aclweb.org/portal/content/acl-code-ethics}{ACL Ethics Policy}. We encourage all authors to include an explicit ethics statement on the broader impact of the work, or other ethical considerations after the conclusion but before the references. The ethics statement will not count toward the page limit (8 pages for long, 4 pages for short papers).
\textbf{Data Privacy}: Although the datasets used in our research are publicly accessible, the utilization of social media conversations for debunking fake news may raise concerns regarding user privacy. To address this issue, we took measures to anonymize all social media posts during the data processing and experiments, ensuring that user information remains invisible and unusable. Additionally, our proposed approach does not require access to any sensitive user information, therefore eliminating the risk of privacy infringement. The collection of social media conversations in the BuzzNews dataset was conducted in compliance with the privacy regulations set by the platform.

\noindent\textbf{Social Implications}: The detection and debunking of fake news can carry significant social and political implications. One critical consideration is the potential impact on the reliability of the system and the possibility of misleading users by mislabeling information as misinformation or vice versa. In light of this concern, we have taken precautions to carefully assess the model we developed and restrict their distribution to the general public. We are committed to designing a responsible policy regarding the dissemination of codes and datasets within research community, and ensure that they are used responsibly in a manner that aligns with ethical standards and societal well-being.

% \section*{Acknowledgements}

\bibliography{anthology,custom}

\begin{thebibliography}{73}
\expandafter\ifx\csname natexlab\endcsname\relax\def\natexlab#1{#1}\fi

\bibitem[{Angelidis and Lapata(2018)}]{angelidis-lapata-2018-multiple}
Stefanos Angelidis and Mirella Lapata. 2018.
\newblock \href {https://doi.org/10.1162/tacl_a_00002} {Multiple instance learning networks for fine-grained sentiment analysis}.
\newblock \emph{Transactions of the Association for Computational Linguistics}, 6:17--31.

\bibitem[{Azevedo et~al.(2021)Azevedo, d’Aquin, Davis, and Zarrouk}]{azevedo2021lux}
Lucas Azevedo, Mathieu d’Aquin, Brian Davis, and Manel Zarrouk. 2021.
\newblock Lux (linguistic aspects under examination): Discourse analysis for automatic fake news classification.
\newblock In \emph{Findings of the Association for Computational Linguistics: ACL-IJCNLP 2021}, pages 41--56.

\bibitem[{Bian et~al.(2020)Bian, Xiao, Xu, Zhao, Huang, Rong, and Huang}]{bian2020rumor}
Tian Bian, Xi~Xiao, Tingyang Xu, Peilin Zhao, Wenbing Huang, Yu~Rong, and Junzhou Huang. 2020.
\newblock Rumor detection on social media with bi-directional graph convolutional networks.
\newblock In \emph{Proceedings of the AAAI conference on artificial intelligence}, volume~34, pages 549--556.

\bibitem[{Castillo et~al.(2011)Castillo, Mendoza, and Poblete}]{castillo2011information}
Carlos Castillo, Marcelo Mendoza, and Barbara Poblete. 2011.
\newblock Information credibility on twitter.
\newblock In \emph{Proceedings of the 20th international conference on World wide web}, pages 675--684.

\bibitem[{Chen et~al.(2022)Chen, Li, Zhang, Sui, Lv, Tun, and Shang}]{chen2022cross}
Yixuan Chen, Dongsheng Li, Peng Zhang, Jie Sui, Qin Lv, Lu~Tun, and Li~Shang. 2022.
\newblock Cross-modal ambiguity learning for multimodal fake news detection.
\newblock In \emph{Proceedings of the ACM Web Conference 2022}.

\bibitem[{Chen et~al.(2017)Chen, Gao, Zhang, Zhao, Liu, and Cai}]{chen2017user}
Zheqian Chen, Ben Gao, Huimin Zhang, Zhou Zhao, Haifeng Liu, and Deng Cai. 2017.
\newblock User personalized satisfaction prediction via multiple instance deep learning.
\newblock In \emph{Proceedings of the 26th International Conference on World Wide Web}, pages 907--915.

\bibitem[{Clark et~al.(2019)Clark, Khandelwal, Levy, and Manning}]{clark2019does}
Kevin Clark, Urvashi Khandelwal, Omer Levy, and Christopher~D Manning. 2019.
\newblock What does bert look at? an analysis of bert’s attention.
\newblock \emph{ACL 2019}, page 276.

\bibitem[{Collobert et~al.(2011)Collobert, Weston, Bottou, Karlen, Kavukcuoglu, and Kuksa}]{collobert2011natural}
Ronan Collobert, Jason Weston, L{\'e}on Bottou, Michael Karlen, Koray Kavukcuoglu, and Pavel Kuksa. 2011.
\newblock Natural language processing (almost) from scratch.
\newblock \emph{Journal of machine learning research}, 12(ARTICLE):2493--2537.

\bibitem[{Dai et~al.(2018)Dai, Xiong, Callan, and Liu}]{dai2018convolutional}
Zhuyun Dai, Chenyan Xiong, Jamie Callan, and Zhiyuan Liu. 2018.
\newblock Convolutional neural networks for soft-matching n-grams in ad-hoc search.
\newblock In \emph{Proceedings of the eleventh ACM international conference on web search and data mining}, pages 126--134.

\bibitem[{Dietterich et~al.(1997)Dietterich, Lathrop, and Lozano-P{\'e}rez}]{dietterich1997solving}
Thomas~G Dietterich, Richard~H Lathrop, and Tom{\'a}s Lozano-P{\'e}rez. 1997.
\newblock Solving the multiple instance problem with axis-parallel rectangles.
\newblock \emph{Artificial intelligence}, 89(1-2):31--71.

\bibitem[{Dun et~al.(2021)Dun, Tu, Chen, Hou, and Yuan}]{dun2021kan}
Yaqian Dun, Kefei Tu, Chen Chen, Chunyan Hou, and Xiaojie Yuan. 2021.
\newblock Kan: Knowledge-aware attention network for fake news detection.
\newblock In \emph{AAAI}.

\bibitem[{Feng et~al.(2012)Feng, Banerjee, and Choi}]{feng2012syntactic}
Song Feng, Ritwik Banerjee, and Yejin Choi. 2012.
\newblock Syntactic stylometry for deception detection.
\newblock In \emph{Proceedings of the 50th Annual Meeting of the Association for Computational Linguistics (Volume 2: Short Papers)}, pages 171--175.

\bibitem[{Foulds and Frank(2010)}]{foulds2010review}
James Foulds and Eibe Frank. 2010.
\newblock A review of multi-instance learning assumptions.
\newblock \emph{The knowledge engineering review}, 25(1):1--25.

\bibitem[{Fung et~al.(2021)Fung, Thomas, Reddy, Polisetty, Ji, Chang, McKeown, Bansal, and Sil}]{fung2021infosurgeon}
Yi~Fung, Christopher Thomas, Revanth~Gangi Reddy, Sandeep Polisetty, Heng Ji, Shih-Fu Chang, Kathleen McKeown, Mohit Bansal, and Avirup Sil. 2021.
\newblock Infosurgeon: Cross-media fine-grained information consistency checking for fake news detection.
\newblock In \emph{Proceedings of the 59th Annual Meeting of the Association for Computational Linguistics and the 11th International Joint Conference on Natural Language Processing (Volume 1: Long Papers)}, pages 1683--1698.

\bibitem[{Glockner et~al.(2022)Glockner, Hou, and Gurevych}]{glockner2022missing}
Max Glockner, Yufang Hou, and Iryna Gurevych. 2022.
\newblock Missing counter-evidence renders nlp fact-checking unrealistic for misinformation.
\newblock In \emph{Proceedings of the 2022 Conference on Empirical Methods in Natural Language Processing}, pages 5916--5936.

\bibitem[{Grail et~al.(2021)Grail, Perez, and Gaussier}]{grail2021globalizing}
Quentin Grail, Julien Perez, and Eric Gaussier. 2021.
\newblock Globalizing bert-based transformer architectures for long document summarization.
\newblock In \emph{Proceedings of the 16th Conference of the European Chapter of the Association for Computational Linguistics: Main Volume}, pages 1792--1810.

\bibitem[{Hu et~al.(2021)Hu, Yang, Zhang, Zhong, Tang, Shi, Duan, and Zhou}]{hu2021compare}
Linmei Hu, Tianchi Yang, Luhao Zhang, Wanjun Zhong, Duyu Tang, Chuan Shi, Nan Duan, and Ming Zhou. 2021.
\newblock Compare to the knowledge: Graph neural fake news detection with external knowledge.
\newblock In \emph{Proceedings of the 59th Annual Meeting of the Association for Computational Linguistics and the 11th International Joint Conference on Natural Language Processing (Volume 1: Long Papers)}, pages 754--763.

\bibitem[{Jin et~al.(2022)Jin, Wang, Yang, Sun, Wang, Liao, and Xie}]{jin2022towards}
Yiqiao Jin, Xiting Wang, Ruichao Yang, Yizhou Sun, Wei Wang, Hao Liao, and Xing Xie. 2022.
\newblock Towards fine-grained reasoning for fake news detection.
\newblock In \emph{Proceedings of the AAAI Conference on Artificial Intelligence}, volume~36, pages 5746--5754.

\bibitem[{Jin et~al.(2016)Jin, Cao, Zhang, Zhou, and Tian}]{jin2016novel}
Zhiwei Jin, Juan Cao, Yongdong Zhang, Jianshe Zhou, and Qi~Tian. 2016.
\newblock Novel visual and statistical image features for microblogs news verification.
\newblock \emph{IEEE transactions on multimedia}, 19(3):598--608.

\bibitem[{Joshi et~al.(2015)Joshi, Kale, Chandel, and Pal}]{joshi2015likert}
Ankur Joshi, Saket Kale, Satish Chandel, and D~Kumar Pal. 2015.
\newblock Likert scale: Explored and explained.
\newblock \emph{British journal of applied science \& technology}, 7(4):396.

\bibitem[{Kaliyar et~al.(2021)Kaliyar, Goswami, and Narang}]{kaliyar2021fakebert}
Rohit~Kumar Kaliyar, Anurag Goswami, and Pratik Narang. 2021.
\newblock Fakebert: Fake news detection in social media with a bert-based deep learning approach.
\newblock \emph{Multimedia Tools and Applications}, 80(8):11765--11788.

\bibitem[{Keerthi and Lin(2003)}]{keerthi2003asymptotic}
S~Sathiya Keerthi and Chih-Jen Lin. 2003.
\newblock Asymptotic behaviors of support vector machines with gaussian kernel.
\newblock \emph{Neural computation}, 15(7):1667--1689.

\bibitem[{Khoo et~al.(2020)Khoo, Chieu, Qian, and Jiang}]{khoo2020interpretable}
Ling Min~Serena Khoo, Hai~Leong Chieu, Zhong Qian, and Jing Jiang. 2020.
\newblock Interpretable rumor detection in microblogs by attending to user interactions.
\newblock In \emph{Proceedings of the AAAI conference on artificial intelligence}, volume~34, pages 8783--8790.

\bibitem[{Kingma and Ba(2015)}]{DBLP:journals/corr/KingmaB14}
Diederik~P. Kingma and Jimmy Ba. 2015.
\newblock \href {http://arxiv.org/abs/1412.6980} {Adam: {A} method for stochastic optimization}.
\newblock In \emph{3rd International Conference on Learning Representations, {ICLR} 2015, San Diego, CA, USA, May 7-9, 2015, Conference Track Proceedings}.

\bibitem[{Kwon et~al.(2013)Kwon, Cha, Jung, Chen, and Wang}]{kwon2013prominent}
Sejeong Kwon, Meeyoung Cha, Kyomin Jung, Wei Chen, and Yajun Wang. 2013.
\newblock Prominent features of rumor propagation in online social media.
\newblock In \emph{2013 IEEE 13th international conference on data mining}, pages 1103--1108. IEEE.

\bibitem[{Lin et~al.(2022)Lin, Ma, Chen, Yang, Cheng, and Guang}]{lin2022detect}
Hongzhan Lin, Jing Ma, Liangliang Chen, Zhiwei Yang, Mingfei Cheng, and Chen Guang. 2022.
\newblock Detect rumors in microblog posts for low-resource domains via adversarial contrastive learning.
\newblock In \emph{Findings of the Association for Computational Linguistics: NAACL 2022}, pages 2543--2556.

\bibitem[{Lin et~al.(2021)Lin, Ma, Cheng, Yang, Chen, and Chen}]{lin2021rumor}
Hongzhan Lin, Jing Ma, Mingfei Cheng, Zhiwei Yang, Liangliang Chen, and Guang Chen. 2021.
\newblock Rumor detection on twitter with claim-guided hierarchical graph attention networks.
\newblock In \emph{Proceedings of the 2021 Conference on Empirical Methods in Natural Language Processing}, pages 10035--10047.

\bibitem[{Lin et~al.(2020)Lin, Moosaei, and Yang}]{lin2020outfitnet}
Yusan Lin, Maryam Moosaei, and Hao Yang. 2020.
\newblock Outfitnet: Fashion outfit recommendation with attention-based multiple instance learning.
\newblock In \emph{Proceedings of The Web Conference 2020}, pages 77--87.

\bibitem[{Liu et~al.(2020)Liu, Xiong, Sun, and Liu}]{liu2020fine}
Zhenghao Liu, Chenyan Xiong, Maosong Sun, and Zhiyuan Liu. 2020.
\newblock Fine-grained fact verification with kernel graph attention network.
\newblock In \emph{Proceedings of the 58th Annual Meeting of the Association for Computational Linguistics}, pages 7342--7351.

\bibitem[{Lu and Li(2020)}]{lu2020gcan}
Yi-Ju Lu and Cheng-Te Li. 2020.
\newblock Gcan: Graph-aware co-attention networks for explainable fake news detection on social media.
\newblock In \emph{Proceedings of the 58th Annual Meeting of the Association for Computational Linguistics}, pages 505--514.

\bibitem[{Luo et~al.(2016)Luo, Chang, and Ban}]{luo2016regression}
Xiong Luo, Xiaohui Chang, and Xiaojuan Ban. 2016.
\newblock Regression and classification using extreme learning machine based on l1-norm and l2-norm.
\newblock \emph{Neurocomputing}, 174(PA):179--186.

\bibitem[{Ma et~al.(2019)Ma, Gao, Joty, and Wong}]{ma2019sentence}
Jing Ma, Wei Gao, Shafiq Joty, and Kam-Fai Wong. 2019.
\newblock Sentence-level evidence embedding for claim verification with hierarchical attention networks.
\newblock In \emph{Proceedings of the 57th Annual Meeting of the Association for Computational Linguistics}, pages 2561--2571.

\bibitem[{Ma et~al.(2020)Ma, Gao, Joty, and Wong}]{ma2020attention}
Jing Ma, Wei Gao, Shafiq Joty, and Kam-Fai Wong. 2020.
\newblock An attention-based rumor detection model with tree-structured recursive neural networks.
\newblock \emph{ACM Transactions on Intelligent Systems and Technology (TIST)}, 11(4):1--28.

\bibitem[{Ma et~al.(2016)Ma, Gao, Mitra, Kwon, Jansen, Wong, and Cha}]{ma2016detecting}
Jing Ma, Wei Gao, Prasenjit Mitra, Sejeong Kwon, Bernard~J Jansen, Kam-Fai Wong, and Meeyoung Cha. 2016.
\newblock Detecting rumors from microblogs with recurrent neural networks.

\bibitem[{Ma et~al.(2015)Ma, Gao, Wei, Lu, and Wong}]{ma2015detect}
Jing Ma, Wei Gao, Zhongyu Wei, Yueming Lu, and Kam-Fai Wong. 2015.
\newblock Detect rumors using time series of social context information on microblogging websites.
\newblock In \emph{Proceedings of the 24th ACM international on conference on information and knowledge management}, pages 1751--1754.

\bibitem[{Ma et~al.(2017)Ma, Gao, and Wong}]{ma2017detect}
Jing Ma, Wei Gao, and Kam-Fai Wong. 2017.
\newblock Detect rumors in microblog posts using propagation structure via kernel learning.
\newblock In \emph{Proceedings of the 55th Annual Meeting of the Association for Computational Linguistics (Volume 1: Long Papers)}, pages 708--717.

\bibitem[{Ma et~al.(2018)Ma, Gao, and Wong}]{ma2018rumor}
Jing Ma, Wei Gao, and Kam-Fai Wong. 2018.
\newblock Rumor detection on twitter with tree-structured recursive neural networks.
\newblock In \emph{Proceedings of the 56th Annual Meeting of the Association for Computational Linguistics (Volume 1: Long Papers)}, pages 1980--1989.

\bibitem[{Mehta et~al.(2022)Mehta, Pacheco, and Goldwasser}]{mehta2022tackling}
Nikhil Mehta, Mar{\'\i}a~Leonor Pacheco, and Dan Goldwasser. 2022.
\newblock Tackling fake news detection by continually improving social context representations using graph neural networks.
\newblock In \emph{Proceedings of the 60th Annual Meeting of the Association for Computational Linguistics (Volume 1: Long Papers)}, pages 1363--1380.

\bibitem[{Min et~al.(2022)Min, Rong, Bian, Xu, Zhao, Huang, and Ananiadou}]{min2022divide}
Erxue Min, Yu~Rong, Yatao Bian, Tingyang Xu, Peilin Zhao, Junzhou Huang, and Sophia Ananiadou. 2022.
\newblock Divide-and-conquer: Post-user interaction network for fake news detection on social media.
\newblock In \emph{Proceedings of the ACM Web Conference 2022}, pages 1148--1158.

\bibitem[{Nguyen et~al.(2020)Nguyen, Vu, and Nguyen}]{nguyen2020BERTweet}
Dat~Quoc Nguyen, Thanh Vu, and Anh-Tuan Nguyen. 2020.
\newblock Bertweet: A pre-trained language model for english tweets.
\newblock In \emph{Proceedings of the 2020 Conference on Empirical Methods in Natural Language Processing: System Demonstrations}.

\bibitem[{Pan et~al.(2018)Pan, Pavlova, Li, Li, Li, and Liu}]{pan2018content}
Jeff~Z Pan, Siyana Pavlova, Chenxi Li, Ningxi Li, Yangmei Li, and Jinshuo Liu. 2018.
\newblock Content based fake news detection using knowledge graphs.
\newblock In \emph{17th International Semantic Web Conference, ISWC 2018}, pages 669--683. Springer Verlag.

\bibitem[{Pappas and Popescu-Belis(2017)}]{pappas2017explicit}
Nikolaos Pappas and Andrei Popescu-Belis. 2017.
\newblock Explicit document modeling through weighted multiple-instance learning.
\newblock \emph{Journal of Artificial Intelligence Research}, 58:591--626.

\bibitem[{Park et~al.(2021)Park, Park, Chin, Kang, and Cha}]{park2021experimental}
Sungkyu Park, Jamie~Yejean Park, Hyojin Chin, Jeong-han Kang, and Meeyoung Cha. 2021.
\newblock An experimental study to understand user experience and perception bias occurred by fact-checking messages.
\newblock In \emph{Proceedings of the Web Conference 2021}.

\bibitem[{Paszke et~al.(2019)Paszke, Gross, Massa, Lerer, Bradbury, Chanan, Killeen, Lin, Gimelshein, Antiga et~al.}]{paszke2019pytorch}
Adam Paszke, Sam Gross, Francisco Massa, Adam Lerer, James Bradbury, Gregory Chanan, Trevor Killeen, Zeming Lin, Natalia Gimelshein, Luca Antiga, et~al. 2019.
\newblock Pytorch: An imperative style, high-performance deep learning library.
\newblock \emph{Advances in neural information processing systems}, 32.

\bibitem[{Popat et~al.(2018)Popat, Mukherjee, Yates, and Weikum}]{popat2018declare}
Kashyap Popat, Subhabrata Mukherjee, Andrew Yates, and Gerhard Weikum. 2018.
\newblock Declare: Debunking fake news and false claims using evidence-aware deep learning.
\newblock In \emph{Proceedings of the 2018 Conference on Empirical Methods in Natural Language Processing}, pages 22--32.

\bibitem[{Potthast et~al.(2018)Potthast, Kiesel, Reinartz, Bevendorff, and Stein}]{potthast2018stylometric}
Martin Potthast, Johannes Kiesel, Kevin Reinartz, Janek Bevendorff, and Benno Stein. 2018.
\newblock A stylometric inquiry into hyperpartisan and fake news.
\newblock In \emph{Proceedings of the 56th Annual Meeting of the Association for Computational Linguistics (Volume 1: Long Papers)}, pages 231--240.

\bibitem[{Reimers and Gurevych(2019)}]{reimers2019sentence}
Nils Reimers and Iryna Gurevych. 2019.
\newblock Sentence-bert: Sentence embeddings using siamese bert-networks.
\newblock In \emph{Proceedings of the 2019 Conference on Empirical Methods in Natural Language Processing and the 9th International Joint Conference on Natural Language Processing (EMNLP-IJCNLP)}.

\bibitem[{Rogers et~al.(2017)Rogers, Zeckhauser, Gino, Norton, and Schweitzer}]{rogers2017artful}
Todd Rogers, Richard Zeckhauser, Francesca Gino, Michael~I Norton, and Maurice~E Schweitzer. 2017.
\newblock Artful paltering: The risks and rewards of using truthful statements to mislead others.
\newblock \emph{Journal of personality and social psychology}, 112(3):456.

\bibitem[{Ruchansky et~al.(2017)Ruchansky, Seo, and Liu}]{ruchansky2017csi}
Natali Ruchansky, Sungyong Seo, and Yan Liu. 2017.
\newblock Csi: A hybrid deep model for fake news detection.
\newblock In \emph{Proceedings of the 2017 ACM on Conference on Information and Knowledge Management}.

\bibitem[{Santia and Williams(2018)}]{santia2018buzzface}
Giovanni~C Santia and Jake~Ryland Williams. 2018.
\newblock Buzzface: A news veracity dataset with facebook user commentary and egos.
\newblock In \emph{Twelfth international AAAI conference on web and social media}.

\bibitem[{Sheng et~al.(2022)Sheng, Cao, Zhang, Li, Wang, and Zhu}]{sheng2022zoom}
Qiang Sheng, Juan Cao, Xueyao Zhang, Rundong Li, Danding Wang, and Yongchun Zhu. 2022.
\newblock Zoom out and observe: News environment perception for fake news detection.
\newblock In \emph{Proceedings of the 60th Annual Meeting of the Association for Computational Linguistics (Volume 1: Long Papers)}, pages 4543--4556.

\bibitem[{Shu et~al.(2019{\natexlab{a}})Shu, Cui, Wang, Lee, and Liu}]{shu2019defend}
Kai Shu, Limeng Cui, Suhang Wang, Dongwon Lee, and Huan Liu. 2019{\natexlab{a}}.
\newblock defend: Explainable fake news detection.
\newblock In \emph{Proceedings of the 25th ACM SIGKDD international conference on knowledge discovery \& data mining}, pages 395--405.

\bibitem[{Shu et~al.(2020)Shu, Mahudeswaran, Wang, Lee, and Liu}]{shu2020fakenewsnet}
Kai Shu, Deepak Mahudeswaran, Suhang Wang, Dongwon Lee, and Huan Liu. 2020.
\newblock Fakenewsnet: A data repository with news content, social context, and spatiotemporal information for studying fake news on social media.
\newblock \emph{Big data}, 8(3):171--188.

\bibitem[{Shu et~al.(2017)Shu, Sliva, Wang, Tang, and Liu}]{shu2017fake}
Kai Shu, Amy Sliva, Suhang Wang, Jiliang Tang, and Huan Liu. 2017.
\newblock Fake news detection on social media: A data mining perspective.
\newblock \emph{ACM SIGKDD explorations newsletter}, 19(1):22--36.

\bibitem[{Shu et~al.(2019{\natexlab{b}})Shu, Zhou, Wang, Zafarani, and Liu}]{shu2019role}
Kai Shu, Xinyi Zhou, Suhang Wang, Reza Zafarani, and Huan Liu. 2019{\natexlab{b}}.
\newblock The role of user profiles for fake news detection.
\newblock In \emph{Proceedings of the 2019 IEEE/ACM international conference on advances in social networks analysis and mining}, pages 436--439.

\bibitem[{Silva et~al.(2021)Silva, Luo, Karunasekera, and Leckie}]{silva2021embracing}
Amila Silva, Ling Luo, Shanika Karunasekera, and Christopher Leckie. 2021.
\newblock Embracing domain differences in fake news: Cross-domain fake news detection using multi-modal data.
\newblock In \emph{Proceedings of the AAAI conference on artificial intelligence}, volume~35, pages 557--565.

\bibitem[{Solovev and Pr{\"o}llochs(2022)}]{solovev2022moral}
Kirill Solovev and Nicolas Pr{\"o}llochs. 2022.
\newblock Moral emotions shape the virality of covid-19 misinformation on social media.
\newblock In \emph{Proceedings of the ACM web conference 2022}, pages 3706--3717.

\bibitem[{Song et~al.(2021)Song, Chen, Chang, Weng, and Shuai}]{song2021adversary}
Yun-Zhu Song, Yi-Syuan Chen, Yi-Ting Chang, Shao-Yu Weng, and Hong-Han Shuai. 2021.
\newblock Adversary-aware rumor detection.
\newblock In \emph{Findings of the Association for Computational Linguistics: ACL-IJCNLP 2021}, pages 1371--1382.

\bibitem[{Tandoc~Jr(2018)}]{tandoc2018five}
Edson~C Tandoc~Jr. 2018.
\newblock Five ways buzzfeed is preserving (or transforming) the journalistic field.
\newblock \emph{Journalism}, 19(2):200--216.

\bibitem[{Wang et~al.(2016)Wang, Ning, Rangwala, and Ramakrishnan}]{wang2016multiple}
Wei Wang, Yue Ning, Huzefa Rangwala, and Naren Ramakrishnan. 2016.
\newblock A multiple instance learning framework for identifying key sentences and detecting events.
\newblock In \emph{Proceedings of the 25th ACM International on Conference on Information and Knowledge Management}, pages 509--518.

\bibitem[{Wang et~al.(2018)Wang, Ma, Jin, Yuan, Xun, Jha, Su, and Gao}]{wang2018eann}
Yaqing Wang, Fenglong Ma, Zhiwei Jin, Ye~Yuan, Guangxu Xun, Kishlay Jha, Lu~Su, and Jing Gao. 2018.
\newblock Eann: Event adversarial neural networks for multi-modal fake news detection.
\newblock In \emph{Proceedings of the 24th acm sigkdd international conference on knowledge discovery \& data mining}, pages 849--857.

\bibitem[{Wang et~al.(2021)Wang, Ma, Wang, Jha, and Gao}]{wang2021multimodal}
Yaqing Wang, Fenglong Ma, Haoyu Wang, Kishlay Jha, and Jing Gao. 2021.
\newblock Multimodal emergent fake news detection via meta neural process networks.
\newblock In \emph{Proceedings of the 27th ACM SIGKDD Conference on Knowledge Discovery \& Data Mining}.

\bibitem[{Wu et~al.(2015)Wu, Yang, and Zhu}]{wu2015false}
Ke~Wu, Song Yang, and Kenny~Q Zhu. 2015.
\newblock False rumors detection on sina weibo by propagation structures.
\newblock In \emph{2015 IEEE 31st international conference on data engineering}, pages 651--662. IEEE.

\bibitem[{Wu et~al.(2019)Wu, Morstatter, Carley, and Liu}]{wu2019misinformation}
Liang Wu, Fred Morstatter, Kathleen~M Carley, and Huan Liu. 2019.
\newblock Misinformation in social media: definition, manipulation, and detection.
\newblock \emph{ACM SIGKDD Explorations Newsletter}, 21(2):80--90.

\bibitem[{Wu et~al.(2021)Wu, Zhan, Zhang, Wang, and Xu}]{wu2021multimodal}
Yang Wu, Pengwei Zhan, Yunjian Zhang, Liming Wang, and Zhen Xu. 2021.
\newblock Multimodal fusion with co-attention networks for fake news detection.
\newblock In \emph{Findings of the association for computational linguistics: ACL-IJCNLP 2021}, pages 2560--2569.

\bibitem[{Xiong et~al.(2017)Xiong, Dai, Callan, Liu, and Power}]{xiong2017end}
Chenyan Xiong, Zhuyun Dai, Jamie Callan, Zhiyuan Liu, and Russell Power. 2017.
\newblock End-to-end neural ad-hoc ranking with kernel pooling.
\newblock In \emph{Proceedings of the 40th International ACM SIGIR conference on research and development in information retrieval}.

\bibitem[{Xu et~al.(2022)Xu, Wu, Liu, Wu, and Wang}]{xu2022evidence}
Weizhi Xu, Junfei Wu, Qiang Liu, Shu Wu, and Liang Wang. 2022.
\newblock Evidence-aware fake news detection with graph neural networks.
\newblock In \emph{Proceedings of the ACM Web Conference 2022}, pages 2501--2510.

\bibitem[{Yang et~al.(2022{\natexlab{a}})Yang, Ma, Lin, and Gao}]{DBLP:conf/sigir/YangMLG22}
Ruichao Yang, Jing Ma, Hongzhan Lin, and Wei Gao. 2022{\natexlab{a}}.
\newblock A weakly supervised propagation model for rumor verification and stance detection with multiple instance learning.
\newblock In \emph{{SIGIR} '22: The 45th International {ACM} {SIGIR} Conference on Research and Development in Information Retrieval, Madrid, Spain, July 11 - 15, 2022}, pages 1761--1772. {ACM}.

\bibitem[{Yang et~al.(2022{\natexlab{b}})Yang, Wang, Jin, Li, Lian, and Xie}]{yang2022reinforcement}
Ruichao Yang, Xiting Wang, Yiqiao Jin, Chaozhuo Li, Jianxun Lian, and Xing Xie. 2022{\natexlab{b}}.
\newblock Reinforcement subgraph reasoning for fake news detection.
\newblock In \emph{Proceedings of the 28th ACM SIGKDD Conference on Knowledge Discovery and Data Mining}, pages 2253--2262.

\bibitem[{Yuan et~al.(2019)Yuan, Ma, Zhou, Han, and Hu}]{yuan2019jointly}
Chunyuan Yuan, Qianwen Ma, Wei Zhou, Jizhong Han, and Songlin Hu. 2019.
\newblock Jointly embedding the local and global relations of heterogeneous graph for rumor detection.
\newblock In \emph{2019 IEEE international conference on data mining (ICDM)}, pages 796--805. IEEE.

\bibitem[{Zheng et~al.(2022)Zheng, Zhang, Guo, Wang, Zang, and Zhang}]{zheng2022mfan}
Jiaqi Zheng, Xi~Zhang, Sanchuan Guo, Quan Wang, Wenyu Zang, and Yongdong Zhang. 2022.
\newblock Mfan: Multi-modal feature-enhanced attention networks for rumor detection.
\newblock IJCAI.

\bibitem[{Zhu et~al.(2022)Zhu, Sheng, Cao, Li, Wang, and Zhuang}]{zhu2022generalizing}
Yongchun Zhu, Qiang Sheng, Juan Cao, Shuokai Li, Danding Wang, and Fuzhen Zhuang. 2022.
\newblock Generalizing to the future: Mitigating entity bias in fake news detection.
\newblock In \emph{Proceedings of the 45th International ACM SIGIR Conference on Research and Development in Information Retrieval}.

\bibitem[{Zubiaga et~al.(2017)Zubiaga, Voss, Procter, Liakata, Wang, and Tsakalidis}]{zubiaga2017towards}
Arkaitz Zubiaga, Alex Voss, Rob Procter, Maria Liakata, Bo~Wang, and Adam Tsakalidis. 2017.
\newblock Towards real-time, country-level location classification of worldwide tweets.
\newblock \emph{IEEE Transactions on Knowledge and Data Engineering}, 29(9):2053--2066.

\end{thebibliography}
\bibliographystyle{acl_natbib}

\appendix

\section{Appendix}
\label{sec:appendix}

\subsection{Detailed Baseline Settings} \label{sec:appendix1}

% Table generated by Excel2LaTeX from sheet 'Sheet2'
\begin{table*}[t!]
  \centering
  \small
  \resizebox{0.98\textwidth}{!}{
    \begin{tabular}{l|c|c}
    \toprule
    \textbf{Method} & \textbf{Article-level} & \textbf{Sentence-level} \\
    \midrule
    \textbf{DeClarE} & \multicolumn{1}{l|}{Title as news content. Posts as evidence.} & \multicolumn{1}{l}{Sentence as claim. Linked posts as evidence.} \\
    \textbf{HAN}   & \multicolumn{1}{l|}{Title as news content. Posts as evidence.} & \multicolumn{1}{l}{Sentence as claim. Linked posts as evidence.} \\
    \textbf{dEFEND} & \multicolumn{1}{l|}{Article as news content. Posts as evidence.} & \multicolumn{1}{l}{Sentence as news content. Linked posts as evidence.} \\
    \textbf{BerTweet} & \multicolumn{1}{l|}{Title as news content. Posts fine-tune model.} & \multicolumn{1}{l}{Sentence as claim. Linked posts fine-tune the model.} \\ % \textcolor{red}{related post feed into the model?}
    \midrule
    \textbf{GCAN}  & \multicolumn{1}{l|}{Title as news content. Users of posts as evidence.} & \multicolumn{1}{l}{Sentence as claim. Users of linked posts as evidence.} \\
    \textbf{Bi-GCN} & \multicolumn{1}{l|}{Title as news content. Posts as evidence} & \multicolumn{1}{l}{Sentence as claim. Linked posts as evidence.} \\
    \textbf{KAN}   & \multicolumn{1}{l|}{Article as news content. Entities from articles and posts.} & \multicolumn{1}{l}{Sentence as claim. Entities from sentence and linked posts.} \\
    \textbf{SureFact} & \multicolumn{1}{l|}{Title as news content. Posts as evidence.} & \multicolumn{1}{c}{--} \\
    \bottomrule
    \end{tabular}%
    }
    \caption{Application of baselines to suit the fake news datasets while keeping their original implementation intact.}
  \label{tab:BaselineImplementation}%
\end{table*}%

Existing fake news detection and rumor detection methods predominately focus on coarse-level classification on the entire article and claim, respectively, while our goals include identifying misinforming sentences within an article at a fine-grained level. When comparing with the baselines that are originally designed to either classify a news article or a claim, the required (and available) inputs may differ from our study. Therefore, we need to specifically customize the data inputs to make the baselines applicable to the article-level and sentence-level detection tasks while maintaining the implementation of baseline models intact. In this section, we will provide more details about baseline models and the information they used. 

\subsubsection{Article-level Task} 

1) \textbf{DeClarE}~\cite{popat2018declare} is designed to classify a claim with relevant news content obtained from external sources as evidence, such as web search results. The claim it used is short and there are many relevant articles providing evidence. In our fake news detection dataset, however, what is available includes a single long-form article which is the target to be checked, and the relevant social conversation trees providing external assistance. Since DeClarE can only accept short claims as input, we use the title of the news article as an input claim and the posts in conversations as evidence. 

%\textcolor{red}{Please describe the following baselines in the similar fashion as DeClarE with sufficient, accurate, and justifiable details, as they're still lack of clarity. If not clear, there is no point to put them as they may confuse reviewers. I strongly suggest you pay attention to the quality of presentation.}

2) \textbf{HAN}~\cite{ma2019sentence} aims similarly to DeClarE to the claim verification task and the provided evidence set is collected from multiple documents relevant to the claim. 
% The claim it used is also short and sentences are from different articles.
In our case, article text is the target to be verified, while HAN assumes a short claim as the target which cannot be fed into HAN directly. So, we use the news title as the input claim and posts in conversations as evidence.

3) \textbf{dEFEND}~\cite{shu2019defend} is a fake news detection model using news article as the target of verification and the related user comments as evidence. This is mostly consistent with our setting. Thus, it does not require any special treatment.

4) \textbf{BerTweet}~\cite{nguyen2020BERTweet} is a pre-trained language model trained on large English posts corpus. It is designed to encode short text. To apply BerTweet for article-level verification, we use the posts in conversation trees to fine-tune the model, and then treat the news title as a claim to be verified because BerTweet cannot accept the long-form article as input.

5) \textbf{GCAN}~\cite{lu2020gcan} aims at debunking rumors only using the corresponding sequence of retweet users without text comments of a source tweet. The source tweet it accepts as a claim is also short. To apply it to our data, we use the news title as source tweet and the post user profiles and propagation structure without post content as evidence.

6) \textbf{Bi-GCN}~\cite{bian2020rumor} utilizes bi-directional Graph Convolutional Network to accommodate top-down and bottom-up post propagation structure to detect rumors taking a short source post as input. Similarly, we use news title as a source post and post propagation structure as evidence.

7) \textbf{KAN}~\cite{dun2021kan} detects fake news by identifying entity mentions in news contents and align them with the entities in the knowledge graph, which are used to learn news-entity co-attentions for better representing news text. While there are news articles in our data, we have only related posts from social media but no knowledge graph. For this issue, we use the social conversions of the article as the source to extract entities as entity contexts of the entities in the article.  

%\ruichao{ the overlap of sentences and related posts as entities to train KAN.}\textcolor{red}{Don't understand. You might identify some entities from article and posts, but how are they correlated? Knowledge graph is used in KAN to help connect related entities, so that their designed attention mechanism make use of the entity relation for better representation of text. I wonder how your implementation can make use of entities.}

8) \textbf{SureFact}~\cite{yang2022reinforcement} groups related posts based on specific topics extracted from news content to implicitly connect news and social media content for fake news detection. It can be directly applied to our datasets.

\subsubsection{Sentence-level Task} 

For misinforming sentence detection, the baselines are deployed by treating a sentence in article as a claim or source post and the conversation trees linked to the sentence (see Section~\ref{sec:linking}) as the source of evidence. In such a setting, most of the baselines can be applied to this sentence-level task in a more straightforward manner. See  Table~\ref{tab:BaselineImplementation} for specific details.

\subsection{Implementation Details}\label{sec:appendix2}
Our model parameters are updated by back-propagation~\cite{collobert2011natural} with Adam~\cite{DBLP:journals/corr/KingmaB14} optimizer. We set the maximum epoch to 100, the dimension of embeddings to 512 for sentences and posts, and empirically initialize the learning rate as 0.001, and the hyperparameter $\lambda$ is set to 0.5 which is validated on a small hold-out dataset. 

As for Gaussian kernels in Equation~\ref{equ:MatchingKernel}, we set $K = 10$. Here one kernel with parameter $\mu_k=1$ and $\sigma_k=0.001$ is designed for exact matching~\cite{dai2018convolutional}. The other kernels' parameter $\sigma_k=0.01$, and their parameter $\mu_k$ is distributed within [-1, 1] evenly. 

The training process is controlled to end when the loss value converges or the maximum epoch number is met.

\subsection{Experiment on Kernel Attention Concentration} \label{sec:apendix3}
We conduct an experiment to compute the entropy values of kernel attention weights used in WSDMS and compare it with dot-product attention used in GCAN, to reflect whether the learned attention weights are more focused or scattered. The lower the entropy, the more focused the attention mechanism~\cite{clark2019does}. The entropy results are given in Table~\ref{tab:AttentionEntropy}.

\begin{table}[htbp]
  \centering
  \small
    \begin{tabular}{lcc}
    \toprule
          & \textbf{kernel} & \textbf{dot-product} \\
    \midrule
    \textbf{Attention Entropy} & 5.11  & 6.03 \\
    \bottomrule
    \end{tabular}%
\caption{Entropy score of kernel attention and dot-product attention.}
% \vspace{-0.2cm}
  \label{tab:AttentionEntropy}%
\end{table}%

We find that kernel attention bears a smaller entropy than the dot-product attention. It suggests that kernel attention has a stronger  ability to be focused on a few more vital posts. This is also the reason why we use kernel attention in our method.

\end{document}